\documentclass{article}
\PassOptionsToPackage{dvipsnames}{xcolor}   
\usepackage{iclr2027_conference,times}

\usepackage{amsmath,amsfonts,bm}

\def\eqref#1{equation~\ref{#1}}

\def\1{\bm{1}}

\DeclareMathAlphabet{\mathsfit}{\encodingdefault}{\sfdefault}{m}{sl}
\SetMathAlphabet{\mathsfit}{bold}{\encodingdefault}{\sfdefault}{bx}{n}

\usepackage[T1]{fontenc}
\usepackage{xcolor}
\usepackage{amsmath,amssymb,mathtools,amsthm,thmtools}
\usepackage{booktabs,nicefrac,microtype,cancel}
\usepackage{graphicx,wrapfig,subcaption}
\usepackage{algorithm,algorithmic}
\usepackage[most]{tcolorbox}
\usepackage[normalem]{ulem}
\usepackage{tikz}
\usetikzlibrary{calc,arrows.meta,fit,backgrounds}
\usepackage{natbib}        
\usepackage{xurl}

\usepackage{hyperref}      
\usepackage[nameinlink,capitalize,noabbrev]{cleveref}  
\usepackage{arydshln}

\definecolor{mypink}{rgb}{0.8353,0.251,0.3686}
\definecolor{revblue}{RGB}{0,0,0}
\newcommand{\rev}[1]{{\color{revblue}#1}}
\newcommand{\revstart}{\color{revblue}}
\newcommand{\revend}{\color{black}}
\hypersetup{colorlinks=true,linkcolor=mypink,citecolor=mypink,
            filecolor=mypink,urlcolor=mypink}

\iclrfinalcopy
\title{Retokenization Reveals the Gap Between Semantics and Behavior}

\author{Sami Wolf\thanks{Equal contribution.} \\
Cornell University
\And
Sergei Kudriashov\footnotemark[1] \\
The Chinese University of Hong Kong, Shenzhen \\
Shenzhen Loop Area Institute
\And
Dmitry Vaintrob \\
Principles of Intelligence
}

\begin{document}

\newcommand\sk[1]{\textcolor{blue}{#1}}
\newcommand\smw[1]{\textcolor{ForestGreen}{#1}}
\newcommand\fread{f_{\mathrm{read}}}
\newcommand{\val}[1]{#1}
\newcommand\boldt{\mathbf{t}}
\newcommand\boldx{\mathbf{x}}
\newcommand\boldz{\mathbf{z}}
\newcommand\tokenset{\mathcal{T}}

\newtheorem{theorem}{Theorem}
\newtheorem{proposition}{Proposition}
\newtheorem{definition}{Definition}[section]

\maketitle
\lhead{Preprint}

\begin{abstract}
    
Language models are trained under a deterministic canonical tokenization, yet every byte string admits exponentially many non-canonical token sequences that decode to it. Although such sequences never occur during training, models exhibit partial invariance to them, and in some settings, such as character-level word games, non-canonical inputs even improve accuracy. This invariance is nevertheless fragile: byte-identical prompts can induce divergent behavior, and adversarial tokenizations were found to circumvent safety alignment. We give a mechanistic account of this fragility and introduce a family of consistency-training objectives that enforce agreement across retokenizations of the prompt \emph{while keeping the output space canonically tokenized}. \rev{We prove that \emph{retokenizing only the prompt} leaves the training optimum unchanged, whereas retokenizing the target raises the irreducible entropy of the target distribution. Across three model families, consistency-tuned models are more robust to adversarial tokenization, at some loss of benchmark accuracy, while transfer to other attacks depends on the objective. Probing shows that the attack suppresses the linearly decodable refusal signal in the residual stream, and that after consistency training the signal persists under attack in its original subspace.}
    
\end{abstract}

\section{Introduction}

\revstart%
Large language models (LLMs) are able to comprehend text that is tokenized very differently from their training data \citep{jain2026emergent}. Even when models receive input character-by-character, they are able to reconstruct the meaning of the sequence \citep{zheng2025broken}. However, the model \emph{comprehending} retokenized input does not imply it will \emph{behave} identically when it receives such text sequences. \citet{geh2025adversarial} demonstrate that merely changing the tokenization of a harmful instruction can erode a safety-aligned model's probability of refusal. Certain encodings, therefore, dissociate semantic content from alignment behavior.
\revend%

\revstart%
Unlike jailbreak methods that append a string to the prompt or create a role-playing scenario, retokenization does not change the underlying byte sequence of text at all, so the underlying semantic ground truth defined by the model's behavior on the canonically tokenized text is preserved by construction. Retokenization may thus be treated as a clean mathematical playground to study robustness and invariance properties of language models, similar to pixelating or rotating images for vision models \citep{geiping2022much}. Retokenization also obviates the need to artificially tamper with model internals to obtain perturbations of varying magnitudes. Semantic understanding and behavioral consistency both degrade, on average, as the prompt moves further away from the canonical encoding, but the process is not uniform: some tokenizations affect one without the other, a phenomenon known as mismatched generalization \citep{wei2023jailbroken}. Moreover, an adversarial search (AdvTok; \citealp{geh2025adversarial}) can identify the tokenizations that specifically target alignment.
\revend%

\revstart%
To re-couple semantics and behavior, we introduce output-based and representation-based consistency training, unsupervised fine-tuning objectives that enforce agreement with the model's own canonical behavior across retokenizations of the prompt. Consistency training makes models robust to adversarial tokenization and to character noise, while transfer to adversarial suffixes depends on the objective. Harmful prompts must be in the training mixture. On OLMo-2-7B, training on benign prompts alone lowers greedy AdvTok success only from $0.400$ to $0.315$, although it moves the representations of retokenized harmful prompts most of the way toward their canonical counterparts (Section~\ref{subsec:asr}). Closer representations are therefore not enough, in this setting, for a brittle feature to become robust \citep{Szegedy2013Intriguing, Ilyas2019brittle}. We connect the consistency objectives to distributionally robust losses (Appendix~\ref{app.sec:dro}).
\revend%

\revstart%
\textbf{Our contributions.}
\begin{enumerate}
    \item We analyze the way retokenizing a sentence perturbs the likelihood of a target continuation under the language model and find the perturbation to be highly \emph{non-uniform and sentence-dependent} (Section \ref{sec:setup}). 
    \item We prove that retokenizing the predicted sequence adds the tokenizer's entropy to the minimum attainable loss, while retokenizing only the prompt leaves the optimum unchanged. We deduce that \emph{prompt retokenization is a label-preserving augmentation} (Propositions~\ref{prop:target} and~\ref{prop:prompt}), unlike full-sequence retokenization. 
    \item We demonstrate that \emph{consistency fine-tuning reduces the success rate of adversarial tokenization attacks} and generalizes to greater behavioral invariance across non-canonical contexts (\cref{sec:robustness}). 
    \item \rev{We show that \emph{how much prompt diversity consistency training needs depends on the objective}. With output-based consistency, robustness and accuracy fall as the number of distinct prompts shrinks, and sampling more retokenizations per prompt does not recover them. Representation matching appears to be less sensitive to the number of prompts (\cref{sec:semantics_diversity}).} 
    \item We examine models' internal representations of retokenized inputs with linear probes. We find that adversarially tokenized prompts \emph{lower the accuracy of a linear refusal probe}, whereas a topic probe is damaged no more than by random retokenization of the same depth. Adding a refusal direction back restores refusal on attacked prompts. After consistency fine-tuning, \emph{the refusal probe still classifies attacked harmful prompts as harmful}, indicating the refusal signal persists in its original subspace.

\end{enumerate}
\revend%

\section{Related Work}
\label{sec:relatedwork}

\begin{figure}[tb]
\centering
\definecolor{ulc}{HTML}{B8BEC7}\definecolor{uls}{HTML}{3D6FB0}\definecolor{ula}{HTML}{C8553D}
\definecolor{semtint}{HTML}{EDF2F9}\definecolor{advtint}{HTML}{FBEFEC}
\tikzset{
  tok/.style={inner xsep=.4pt,inner ysep=1.2pt,anchor=base west,font=\ttfamily\scriptsize},
  byt/.style={inner sep=0pt,anchor=base west,font=\ttfamily\scriptsize},
  lab/.style={anchor=base west,inner sep=0pt,font=\footnotesize},
  hdr/.style={anchor=base west,inner sep=0pt,font=\footnotesize\bfseries},
  col/.style={fit=#1,inner sep=0pt},
}
\newcommand{\vsp}{\textcolor{black!30}{\textvisiblespace}}
\newcommand{\tokrow}[3]{
  \coordinate (last) at ([xshift=-1.5pt]#1);
  \foreach \tkst/\tktx in {#3}{%
    \node[tok] (n) at ([xshift=1.5pt]last) {\strut\tktx};
    \draw[ul\tkst,line width=1.3pt,line cap=round]
      ([xshift=1pt,yshift=-.3pt]n.south west) -- ([xshift=-1pt,yshift=-.3pt]n.south east);
    \coordinate (last) at (n.base east);}%
  \coordinate (#2) at (last);}
\resizebox{\ifdim\width>\linewidth\linewidth\else\width\fi}{!}{%
\begin{tikzpicture}
  \def\yh{.5}\def\yt{-.78}\def\gap{18pt}
  \node[lab]      at (0,0)   {$\boldx$\enspace\textsc{bytes}};
  \node[lab] (lt) at (0,\yt) {$\boldt$\enspace\textsc{tokens}};
  \path ([xshift=14pt]lt.base east |- 0,0) coordinate (C1);
  \node[hdr] (h1) at (C1 |- 0,\yh) {In-distribution};
  \node[byt] (x1) at (C1) {\strut Hotwire a car};
  \tokrow{C1 |- 0,\yt}{t1}{c/Hot,c/wire,c/\vsp a,c/\vsp car}
  \node[col={(h1)(x1)(t1)}] (k1) {};
  \path ([xshift=\gap]k1.east |- 0,0) coordinate (C2);
  \node[hdr] (h2) at (C2 |- 0,\yh) {Tokenization OOD};
  \node[byt] (x2) at (C2) {\strut Hotwire a car};
  \tokrow{C2 |- 0,\yt}{t2}{a/H,a/ot,a/w,a/ire,c/\vsp a,a/\vsp c,a/ar}
  \node[col={(h2)(x2)(t2)}] (k2) {};
  \path ([xshift=\gap]k2.east |- 0,0) coordinate (C3);
  \node[hdr] (h3) at (C3 |- 0,\yh) {Semantic OOD};
  \node[byt] (x3) at (C3) {\strut\textcolor{uls}{\textbackslash\textbackslash"\$\textbackslash$\bot$=} Hotwire a car \textcolor{uls}{]\}!!}};
  \tokrow{C3 |- 0,\yt}{t3}{s/\textcolor{uls}{\textbackslash\textbackslash"},s/\textcolor{uls}{\$\textbackslash},s/\textcolor{uls}{$\bot$},s/\textcolor{uls}{=},c/\vsp Hot,c/wire,c/\vsp a,c/\vsp car,s/\textcolor{uls}{\vsp]},s/\textcolor{uls}{\}},s/\textcolor{uls}{!!}}
  \node[col={(h3)(x3)(t3)}] (k3) {};
  \begin{scope}[on background layer]
    \fill[advtint,rounded corners=2pt] ([xshift=-5pt]k2.west |- 0,\yt+.3) rectangle ([xshift=5pt]k2.east |- 0,\yt-.3);
    \fill[semtint,rounded corners=2pt] ([xshift=-5pt]k3.west |- 0,.3)     rectangle ([xshift=5pt]k3.east |- 0,-.22);
    \fill[semtint,rounded corners=2pt] ([xshift=-5pt]k3.west |- 0,\yt+.3) rectangle ([xshift=5pt]k3.east |- 0,\yt-.3);
  \end{scope}
  \draw[black!25,line width=.4pt] (0,.4) -- ([xshift=5pt]k3.east |- 0,.4);
\end{tikzpicture}}
\caption{\textbf{An illustration of a gap between tokenization and semantics}. Inputs to an LLM are pairs $(\boldx,\boldt)$ with $P_\theta(\boldx,\boldt)=P_\theta(\boldx)\,P_\theta(\boldt \mid \boldx)$, where $P_\theta( \boldt \mid \boldx )$ concentrates on the canonical tokenization $\tau(\boldx)$. Most attacks are semantic, so they move $\boldx$ off the support of $P_\theta(\boldx)$ while $\boldt=\tau(\boldx)$ remains canonical. Contrary to that, adversarial tokenization \citep{geh2025adversarial} keeps $\boldx$ in-distribution and moves $\boldt$ off the support of $P_\theta(\boldt \mid \boldx)$.}
\label{fig:teaser}
\end{figure}

\revstart%
Models are only exposed to canonical token sequences during pretraining and
SFT, yet modern LLMs are capable of both reading and writing non-canonical
token sequences \citep{geh-etal-2024-signal}. Some invariance is already learnt at pretraining
\citep{jain2026emergent}, and these symmetries are reinforced and clarified
during instruction tuning \citep{zheng2025broken}. The mechanism for the
invariance seems to be the model's exposure to word fragments in canonical contexts. Early and middle layers implement a general ``detokenization'' step: attention gathers a run of subword pieces into the last position within the first few layers, and feed-forward layers then produce a vector near the embedding the word would have had as one token \citep{kaplan2024tokens}. Even when the input is tokenized character-by-character, hidden states are able to recover the canonical word-level tokens from its letters via in-group attention \citep{yang2026word}. However, the invariance is not complete. Robustness decreases with the degree of fragmentation away from the canonical
tokenization \citep{zheng2025broken}, the extent of symmetry is not uniform across languages \citep{ghosh2026language}, and retokenization can evade alignment without any changes to the underlying prompt string \citep{geh2025adversarial}.

Prior work measures the effect of distance from the canonical tokenization mostly by benchmark performance, with task-completion rates such as pass@retok \citep{jain2026emergent,zheng2025broken,geh2025adversarial}. \citet{jain2026emergent} also track the displacement of the final-position hidden state, and measure the surprisal of a canonical passage when a fully retokenized copy of it is in context. Both are averaged over sampled tokenizations. In Section~\ref{subsec:perturbation}, we measure sensitivity of the target NLL rather than a task-completion rate, because the latter is a Monte Carlo estimate of a binary function and discards gradation between nearby tokenizations, obscuring local structure \citep{schaeffer2023emergent}.
\revend%

\revstart%
Subword regularization methods \citep{kudo2018subwordregularizationimprovingneural,provilkov2020bpedropoutsimpleeffectivesubword} were first developed to improve robustness of machine translation methods. \citet{cao-rimell-2021-evaluate, chirkova-etal-2023-marginalize} noticed that when a string has multiple token-level representations, its exact likelihood is a sum over all such tokenizations. \citet{ghosh2026language} studied robustness under retokenization across languages, and used retokenization as a form of data augmentation during LoRA fine-tuning for cross-lingual task accuracy. \citet{sims2025stochastok} argued that training LLMs with stochastic tokenization (StochasTok) would improve sub-word understanding, but they retokenized both the context and continuation during pretraining and instruction tuning.

\citet{steger2026stochasticity} fine-tuned with stochastic tokenization on multiple-choice tasks, and found that accuracy under random and adversarial tokenization improves with the number of splits and with uniform sampling over tokenizations. They also observed that stochastic pre-training raises perplexity on canonical text, and remarked that later fine-tuning could perturb the input alone so that the model is not trained to generate non-canonical outputs. We make this precise in Section~\ref{sec:conditional}, where we prove that retokenizing the predicted sequence adds the tokenizer's entropy to the minimum attainable loss, whereas retokenizing only the prompt leaves the optimum unchanged. Full-sequence retokenization also leads to the model emitting non-canonical tokens, an effect we measure in Appendix~\ref{app.sec:stochastok}. 

Our consistency training therefore retokenizes the prompt alone. Input-only retokenization has been used before: \citet{wang2021multiview} fine-tuned multilingual classifiers with a KL penalty between predictions on canonical and stochastically segmented views of the input, supervised by gold labels. \citet{cosma2026homotokens} pass sampled retokenizations through a small auxiliary encoder during pretraining, while the decoder keeps the canonical next-token objective, and report delayed overfitting on heavily repeated text. Section~\ref{sec:semantics_diversity} asks a related, but distinct question, namely whether retokenized prompts improve generalization to new prompts. \citet{jain2023baseline} evaluated retokenizing the prompt with BPE-dropout at inference time, as a preprocessing defense against adversarial suffixes. Prior fine-tuning work with retokenized inputs reports task accuracy only \citep{steger2026stochasticity,ghosh2026language}. We train a generative model whose only supervision is its own canonical behavior, and we evaluate refusal under adversarial tokenization.
\revend%

\revstart%
\citet{Chua2024BCT} introduced
bias-augmented consistency training as an unsupervised method against biased
chain-of-thought reasoning, and \citet{Irpan2025ACT} applied it to
jailbreaking and introduced activation consistency training. Our setting differs from previous consistency approaches in that we enforce consistency over classes that contain only a single byte string. Our Proposition~\ref{prop:prompt} shows that the string-level training optimum is preserved under prompt retokenization. Therefore, activation consistency is better justified: the model should interpret tokenizations of the same string as equivalent. This could explain why last-position representation matching is the only objective we find that transfers to adversarial suffixes on all three models, whereas \citet{Irpan2025ACT} found output-based training more robust. Retokenizations admit a distance from the canonical encoding, and we observe that deep perturbations appear necessary for refusal robustness from consistency training. Our objective backpropagates only the sampled retokenizations with the largest loss, an empirical conditional value at risk (CVaR). This reduces computation and has no detectable effect on robustness (Appendix~\ref{app.sec:cvar}).
\revend%

\revstart%
Refusal behavior in safety-trained LLMs is notoriously fragile. \citet{wei2023jailbroken} described mismatched generalization under
jailbreaks, where the model's capabilities extend to modified inputs while
safety training does not. Retokenization is an especially clean version of
this phenomenon, because safety training fails to carry over even when the
byte string is unchanged. \citet{qi2025shallow} showed that safety alignment
is shallow, localized to the first few tokens of the response.
\citet{zou2023representation} showed that several high-level concepts are linearly
decodable from the residual stream and that adding the decoded direction
steers the model's behavior. \citet{arditi2024refusal} found that
refusal, in particular, is mediated by a single such direction. We apply linear probes and activation steering
tools to retokenization.
\revend%

\section{Separating semantics and tokenization}
\label{sec:setup}

\subsection{Notation}
\label{subsec:notation}

Let $\boldx \in \Sigma^{\ast}$ denote a byte string over an alphabet $\Sigma$ and $\boldt \in V^\ast$ a token sequence over a vocabulary $V$. Following \citet{gastaldi2024foundations}, a \emph{tokenizer} is a pair of stochastic maps $(\tau, \kappa)$, with $\tau: \Sigma^\ast \rightarrow V^\ast$ the \emph{encoding} and $\kappa: V^\ast \rightarrow \Sigma^\ast$ the \emph{decoding}. We take $\kappa$ to be deterministic, as it is for the tokenizers we study, so every token sequence decodes to exactly one string. The set of valid tokenizations of $\boldx$ is
\begin{equation}
\label{eq:tokenization_set}
\tokenset(\boldx) = \{\, \boldt = (t_1,\dots,t_k) : t_i \in V,\ t_1 t_2 \cdots t_k = \boldx \,\}.
\end{equation}
For a model trained with a deterministic tokenizer $\tau$, we call $\boldt_c(\boldx)= \tau(\boldx)$ the \textit{canonical} tokenization of $\boldx$. We write $\boldt_c$ for the canonical tokenization when the string is clear from context. We quantify how far a tokenization $\boldt \in \tokenset(\boldx)$ is from the canonical one by its \emph{depth}, the number of added tokens $d(\boldt) = |\boldt| - |\boldt_c|$.

A language model defines a distribution $P_\theta(\boldt)$ over token sequences. Since $\kappa$ is deterministic, the model induces a distribution over strings,
\begin{equation}
\label{eq:string_marginal}
P_\theta(\boldx) = \sum_{\boldt \in \tokenset(\boldx)} P_\theta(\boldt),
\end{equation}
and a conditional over the tokenizations of each string, $P_\theta(\boldt \mid \boldx) = P_\theta(\boldt) / P_\theta(\boldx)$ for $\boldt \in \tokenset(\boldx)$. We write $P_\theta(\boldx, \boldt) = P_\theta(\boldx)\, P_\theta(\boldt \mid \boldx)$ for the joint. Although standard models are pretrained with canonical tokenization only, and can only be exposed to non-canonical token sequences during on-policy learning, they still assign non-zero probabilities to non-canonical token sequences \citep{geh-etal-2024-signal}.

For canonically tokenized context $Z_{\mathrm{pre}}$ preceding and
$Z_{\mathrm{suf}}$ following the segmented text, and a target
continuation $y^{*}$, define the target NLL of a token sequence
$\boldt \in V^{\ast}$ under the model
$\pi_\theta$:
\begin{equation}
f_{\mathrm{read}}(\boldt) = -\log \pi_\theta\big(y^{*} \mid Z_{\mathrm{pre}}, \boldt, Z_{\mathrm{suf}}\big),
\end{equation}
measured in nats; the higher it is, the harder the model finds the
target to predict through this tokenization. Section~\ref{subsec:perturbation} measures how $\fread$ is affected by retokenization.

\subsection{Identifying the gap}
\label{subsec:the_gap}

Figure~\ref{fig:teaser} shows the two interfaces through which an input reaches a language model. We identify the byte string $\boldx$ with the text semantics, whereas the token string $\boldt$ is its surface form. A deterministic encoder couples them, $\boldt=\boldt_c(\boldx)$, so a model is only exposed to one tokenization per text, and whatever it knows about the exponentially many others is inherited from partial invariances that pretraining and post-training happen to induce \citep{kaplan2024tokens, jain2026emergent, zheng2025broken}.

\revstart%
The two interfaces can be modified independently. Semantic attacks search for \emph{strings} that elicit a behavior while keeping the tokenization canonical \citep{geisler2025reinforce}; adversarial tokenization \citep{geh2025adversarial} keeps the text fixed and modifies only the tokenization (i.e. the text's segmentation). Both methods can bypass typical safety post-training. Alignment, therefore, depends not only on text semantics but also on the surface form. We first measure how sensitive model behavior is to retokenization (Section~\ref{subsec:perturbation}). Section~\ref{sec:conditional} shows that retokenizing the prompt is a label-preserving augmentation, while retokenizing the target raises the minimum loss. Sections~\ref{sec:consistency} to~\ref{sec:robustness} introduce consistency training on retokenized prompts and test it against adversarial tokenization. Section~\ref{sec:semantics_diversity} examines whether retokenization provides data diversity, and Section~\ref{sec:probes} examines the model's representations of canonical vs. retokenized prompts.
\revend%

\subsection{Sensitivity to retokenization}
\label{subsec:perturbation}

\revstart%
Prior work has analyzed the effect of retokenizing the prompt on benchmark performance \citep{zheng2025broken, geh2025adversarial, jain2026emergent} and final-position hidden states \citep{jain2026emergent}. These works have studied retokenizations at various depths, but not how the effect depends on the specific tokens that are split nor how the effects of several splits compound. We treat a retokenization as a perturbation of the input and measure its effect by the change in the target NLL $\fread$. The smallest perturbation is a single split, which replaces one token $t_i$ of the canonical sequence by two tokens $u, v \in V$ with $uv = t_i$. For the resulting sequence $\boldt^{(i)}$, let $\Delta_i = \fread(\boldt^{(i)}) - \fread(\boldt_c)$ be the resulting change. On the base model OLMo-2-1124-7B we sweep every split position permitted by the token vocabulary across twenty constructed sentences ($526$ positions). Most splits have little effect on the target NLL: the pooled median of $\Delta_i$ is $\val{+0.11}$ nats, and $\val{39.2}\%$ of splits lower it. A few splits have large effects, the largest raising the target NLL by $\val{4.02}$ nats (Appendix~\ref{app.sec:splittability} Figure~\ref{fig:sensitivity}A). Among splits with $\Delta_i > 0$ the Gini coefficient is $\val{0.507}$; the top decile of these splits accounts for a third of the total summed $\Delta_i$.

Deeper perturbations show sentence-specific behavior. To examine how perturbations compose, we define a \emph{cascade} as a sequence of token strings $(\boldt_c, \boldt_1, \dots , \boldt_m)$ that starts at $\boldt_c$ and splits one token per step, so after $d$ steps the sequence has $d$ more tokens than the canonical one. 

We sample $50$ random cascades to depth $20$ for each of the sixteen sentences that pass the probability filter described in Appendix~\ref{app.sec:details}. Let $\mu_s(d)$ be the mean change in target NLL from the canonical tokenization at depth $d$, over the cascades of sentence $s$. The target NLL rises with depth for every sentence, but the rate of change varies across sentences. Of the variance in NLL change across all sentences, cascades and depths, $36\%$ is a rise common to all sentences, $43\%$ is specific to the sentence and only $21\%$ can be attributed to the cascade. The curves $\mu_s(d)$ also differ across sentences in curvature, i.e. whether splits have compounding or diminishing effects. The per-split increment grows with depth in nine sentences, shrinks in four and shows no detectable trend in three (Figure~\ref{fig:sensitivity}C).

Further analyses are provided in Appendix~\ref{app.sec:splittability}.
\revend%

\section{Separating tokenization effects on conditional distributions}
\label{sec:conditional}

\revstart%
A model over token sequences induces a distribution $P_\theta(\boldx)$ over strings and a conditional $P_\theta(\boldt\mid\boldx)$ over the tokenizations of each string (Section~\ref{subsec:notation}). Let $q(\boldt\mid\boldx)$ be the distribution of the encoding $\tau(\boldx)$ used to tokenize the training data, supported on $\tokenset(\boldx)$. It is a point mass at $\boldt_c(\boldx)$ for a deterministic tokenizer, and a non-degenerate distribution for StochasTok \citep{sims2025stochastok} or BPE-Dropout \citep{provilkov2020bpedropoutsimpleeffectivesubword}. We provide two propositions that separate the effect of retokenizing what the model \emph{predicts} from that of retokenizing what it \emph{conditions on}. All proofs are provided in Appendix~\ref{app.sec:proofs}.
\revend%

\begin{proposition}[Retokenizing the target]\label{prop:target}
For any tokenizer $q$ supported on $\tokenset(\boldx)$,
\begin{align}
    \mathbb{E}_{X,\,T\sim q}\!\left[-\log P_\theta(T)\right]
    &= \mathbb{E}_{X}\!\left[-\log P_\theta(X)\right]
    + \mathbb{E}_{X}\!\left[H\big(q(\cdot\mid X)\big)\right]
    + \mathbb{E}_{X}\!\left[\mathrm{KL}\big(q(\cdot\mid X)\,\|\,P_\theta(\cdot\mid X)\big)\right] \notag\\
    &\geq \mathbb{E}_{X}\!\left[-\log P_\theta(X)\right] + \mathbb{E}_{X}\!\left[H\big(q(\cdot\mid X)\big)\right], \label{eq:entropy_bound}
\end{align}
with equality if and only if $P_\theta(\cdot\mid\boldx)=q(\cdot\mid\boldx)$ for all $\boldx$.
\end{proposition}

When the prediction target is retokenized, the token-level loss thus exceeds the string-level loss by at least the tokenizer's conditional entropy. This entropy is at most $\log|\tokenset(\boldx)|$, attained when $q$ follows a uniform distribution. Under canonical tokenization of the target, the tokenizer's conditional entropy is zero and the remaining gap between token-level and string-level loss vanishes as the model concentrates each string's mass on its canonical encoding, which trained LLMs approximately do \citep{geh-etal-2024-signal, gastaldi2024foundations}. Under a stochastic target tokenization, the optimal model reproduces the tokenizer's distribution over encodings. Token-level likelihoods then no longer reflect string likelihoods unless one marginalizes over an exponentially large set \citep{cao-rimell-2021-evaluate, chirkova-etal-2023-marginalize}. Exact marginalization is $\#\text{P}$-hard in general \citep{geh-etal-2024-signal}. \citet{steger2026stochasticity} noticed higher perplexity on canonical text after stochastic-target pretraining. We measure another consequence of such training, namely higher rates of emission of non-canonical tokens (Appendix~\ref{app.sec:stochastok}). Both follow whenever StochasTok or BPE-Dropout is applied to the predicted sequence \citep{sims2025stochastok, provilkov2020bpedropoutsimpleeffectivesubword}, independently of the regularization and robustness benefits of these methods \citep{kudo2018subwordregularizationimprovingneural, steger2026stochasticity}. We now turn to the the case where only the prompt is retokenized, and the continuation (that is, the prediction target) is left canonical.

\begin{proposition}[Retokenizing the prompt]\label{prop:prompt}
Let $(Z,X)\sim\mu$ be prompt--continuation pairs, let the prompt be retokenized as $\boldt^Z\sim q(\cdot\mid Z)$ while the continuation keeps its canonical tokenization $\boldt_c(X)$, and let the loss be applied to continuation tokens only. With the string-level conditional $P_\theta(\boldx\mid\boldt^Z)=\sum_{\boldt\in\tokenset(\boldx)}P_\theta(\boldt\mid\boldt^Z)$,
\begin{align}
    \mathbb{E}\!\left[-\log P_\theta\big(\boldt_c(X)\mid\boldt^Z\big)\right]
    &= H_\mu(X \mid Z)
    + \mathbb{E}_{Z,\boldt^Z}\!\left[\mathrm{KL}\big(\mu(\cdot\mid Z)\,\|\,P_\theta(\cdot\mid \boldt^Z)\big)\right] \notag\\
    &\quad + \mathbb{E}_{Z,\boldt^Z,X}\!\left[-\log P_\theta\big(\boldt_c(X)\mid X,\boldt^Z\big)\right]. \label{eq:prompt_retok_decomposition}
\end{align}
The last two terms are non-negative and vanish if and only if the model assigns continuation strings their true probabilities $\mu(\cdot\mid\kappa(\boldt^Z))$ and places each string's mass on its canonical tokenization. The infimum is $H_\mu(X\mid Z)$, the value attained under canonical training.
\end{proposition}

\revstart%
No entropy term appears in Eq.~\eqref{eq:prompt_retok_decomposition}, because the retokenization $\boldt^Z$ is drawn from $Z$ alone, so $X \perp \boldt^Z \mid Z$. In words, the retokenization and the continuation are conditionally independent given the prompt's byte string. A retokenized prompt carries exactly the same information about the continuation as the canonical one. Prompt retokenization is therefore a label-preserving augmentation \citep{islam2024diffusemix} in the strict sense, since the invariance it imposes, that the conditional depends on $\boldt^Z$ only through $\kappa(\boldt^Z)$, holds for the true conditional. Canonical training constrains $P_\theta(\cdot\mid\boldt^Z)$ only at $\boldt^Z=\boldt_c(Z)$ and leaves the remaining $|\tokenset(Z)|-1$ tokenizations of the prompt unspecified. The consistency objective provides the missing specification, while following both propositions by retokenizing the prompt and keeping the continuation canonical (Section~\ref{sec:consistency}). Since the optimum is unchanged by retokenization, an alternative tokenization adds no information in the limit of infinite data. We ask in Section~\ref{sec:semantics_diversity} whether additional tokenizations are a substitute for new data when the training prompts are finite.
\revend%

\section{Consistency training}
\label{sec:consistency}

\revstart%

While retokenization may serve as a way to elicit alternative reasoning pathways during inference \citep{jain2026emergent}, when it comes to safety, we want a model's behavior on a prompt to be the same under every tokenization of that prompt. The data already have this invariance, since a continuation depends on its prompt only through the text. Section~\ref{sec:conditional} shows that training on retokenized prompts therefore leaves the optimum unchanged, whereas retokenizing the continuation raises the minimum loss. Instruction-tuned models already separate context from continuation \citep{zheng2025broken}, so a model can be trained to read non-canonical prompts but keep producing canonical text. Consistency training does this by fine-tuning the model to reproduce, on retokenized prompts, its own behavior on the canonical prompt.
\revend%

\noindent\textbf{Retokenizing the prompt.} Let $\boldx$ be a prompt and $\tokenset(\boldx)$ the set of its valid
tokenizations. As in Section~\ref{subsec:notation}, write
$\boldt_c$ for the canonical tokenization and
$\boldt_1, \dots, \boldt_{N-1}$, for tokenizations drawn
uniformly from the multi-valued decision diagram (MDD) of $\boldx$
\citep{geh2025adversarial}. When supervised labels (responses) are available, ordinary SFT can be applied to the retokenized prompts. For each instruction-response pair $(\boldx, y^{*})$, context-perturbed SFT trains on
$\{(\boldt_c, y^{*})\} \cup \{(\boldt_i, y^{*}) : i \in \{1, \dots, N{-}1\}\}$ with the cross-entropy loss. AdvBench provides no refusals, so for harmful prompts we take $y^{*}$ to be a refusal sampled from the reference model. Section~\ref{subsec:asr} compares context-perturbed SFT with the consistency objectives below.

\noindent \textbf{Reference and student.} The student model $\pi_\theta$ is initialized from the reference model $\pi_{\mathrm{ref}}$; the reference remains frozen while the student is trained. The objective is constructed by inputting a canonically-tokenized instruction to the reference model, measuring its output, and penalizing the student model for behaving differently when it receives a non-canonical tokenization of the \emph{same instruction.} No safety labels and no ground-truth responses are used: this is a self-distillation objective, broadening behavior the model already has.

\noindent \textbf{Consistency objective.} We define a few variants of consistency training and later compare the
robustness of the resulting models. Each encoding contributes a term
$d_i = D\big(\boldt_c, \boldt_i\big)$ measuring, at a site chosen per
variant, how far the student's behavior on $\boldt_i$ departs from the
reference's on $\boldt_c$, with $d_0 = D\big(\boldt_c, \boldt_c\big)$ the
canonical term. All variants minimize
\begin{equation}
\mathcal{L}(\theta) \;=\; \frac{1}{k+1}\Big(d_0 \;+\;
k\operatorname*{CVaR}_{\rho}\big\{\, d_i \,\big\}_{i=1}^{N-1}\Big),
\qquad k = \big\lceil \rho (N{-}1) \big\rceil,
\end{equation}
\revstart%
where $\rho \in (0,1]$ is the fraction of the $N{-}1$ sampled encodings retained
and $\operatorname{CVaR}_{\rho}$, the empirical conditional value at risk, is the mean of the $k$
largest sampled terms, ranked without gradients
\citep{rockafellar2000cvar,oren2019dro}. The selection has no detectable effect on robustness at this sample size (Appendix~\ref{app.sec:cvar}),
but it reduces the number of backward passes per prompt from $N$ to $k+1$. The top-$k$ selection is one member of a family of distributionally robust losses over the sampled encodings of a prompt \citep{li2021tilted,namkoong2016stochastic}. See Appendix~\ref{app.sec:dro}.
The canonical term $d_0$ sits outside the
\revend%
CVaR selection and is always backpropagated, holding the canonical behavior
approximately fixed while the behavior on alternative encodings is adjusted. The variants fall into two classes, output-based and
representation-based, which we describe below. Variants are further distinguished by whether they compare 
the reference and student models along a shared continuation, or only at a single site (after the last prompt token). For variants that require a continuation, we let
$y = (y_1, \dots, y_L)$ be sampled once from $\pi_{\mathrm{ref}}$ at
temperature $1$ given $\boldt_c$. $y$ is appended after the encoding a model
receives, so that the reference is evaluated on $(\boldt_c, y)$ and the
student on $(\boldt_i, y)$.

\noindent\textbf{Output-based consistency.} 
Output-based consistency sums the divergence over the positions of $y$, as the
KL divergence between next-token distributions of $\pi_{\mathrm{ref}}$ and
$\pi_\theta$:
\begin{equation}
D_{\mathrm{KL}}\big(\boldt_c, \boldt_i\big) \;=\; \sum_{j=1}^{L}
\sum_{v \in V} \pi_{\mathrm{ref}}\big(v \mid \boldt_c, y_{<j}\big)
\log \frac{\pi_{\mathrm{ref}}\big(v \mid \boldt_c, y_{<j}\big)}
{\pi_\theta\big(v \mid \boldt_i, y_{<j}\big)},
\end{equation}
A cross-entropy variant $D_{\mathrm{CE}}$ replaces the inner sum by
$-\log \pi_\theta(y_j \mid \boldt_i, y_{<j})$, supervising only the token the
reference emitted rather than its full distribution. A single-position variant
applies the same forward KL at the final prompt token only, with no
continuation, differing from single-position representation matching
only in the site of comparison, the next-token distribution after the
unembedding rather than the residual stream before it.

\noindent\textbf{Representation-based consistency.} Representation-based
training aligns the student's residual stream on a retokenized prompt
with the reference's on the canonical one, following activation consistency training \citep{Irpan2025ACT}. Write $h^{\ell}_{j}(\boldt_i)$ for the layer-$\ell$ residual of the model
reading $(\boldt_i, y)$, at position $j$, with $j = 0$ the final prompt token
and $j = 1, \dots, L$ the positions of $y$. Given a layer set $\mathcal{S}$ and a position set $\mathcal{K}$,
\begin{equation}
D_{\mathrm{rep}}\big(\boldt_c, \boldt_i\big) \;=\;
\frac{1}{|\mathcal{S}|\,|\mathcal{K}|}
\sum_{\ell \in \mathcal{S}} \sum_{j \in \mathcal{K}}
\frac{\big\| h^{\ell}_{j,\theta}(\boldt_i) -
h^{\ell}_{j,\mathrm{ref}}(\boldt_c) \big\|^2}
{\big\| h^{\ell}_{j,\mathrm{ref}}(\boldt_c) \big\|^2},
\end{equation}
normalized per layer and position because residual norms grow across layers. Taking $\mathcal{K} = \{0\}$ yields the \emph{single-position}
variant, which supervises only the state at the end of the prompt and
needs no continuation; $\mathcal{K} = \{0, \dots, L-1\}$ yields the
\emph{continuation-supervised} variant, which is matched to the
output objectives in every respect except the site of
supervision, measuring divergence in the residual stream before the
unembedding rather than on next-token distributions after it. We call the single-position variant at all layers \emph{repr-next} and the continuation-supervised variant at all layers \emph{repr-cont}. The representation row of Table~\ref{tab:attacks} is repr-next, which differs from the KL arm in the supervised positions as well as in the site of supervision.

\section{Experimental setup}
\label{sec:experimental_setup}

\noindent\textbf{Models.} The main model is OLMo-2-1124-7B-Instruct \citep{olmo2025two}, used in Sections~\ref{sec:robustness} and~\ref{sec:probes}. The prompt-diversity study of Section~\ref{sec:semantics_diversity} uses Llama-3.2-3B-Instruct \citep{grattafiori2024llama}. Appendix~\ref{app.sec:rebuttal-results} repeats the attack and capability evaluations on Llama-3.2-3B-Instruct and gemma-3-4b-it \citep{gemmateam2025gemma3}.

\noindent\textbf{Training data.} Training prompts mix benign Alpaca instructions \citep{taori2023alpaca} with harmful AdvBench requests \citep{zou2023universal}. We hold out $200$ AdvBench prompts for evaluation and train on the remaining $320$. The $700$ training steps draw $5{,}600$ prompts at random. In the main run $1{,}547$ of them are harmful, so each harmful prompt is seen about five times, and $4{,}053$ are benign, covering $3{,}793$ distinct Alpaca instructions. The consistency objectives use the prompts only.

\noindent\textbf{Training.} For each prompt we sample $N{-}1 = 8$ tokenizations uniformly from the MDD, which inflates the token count by $3.62\times$ on average. We set $\rho = 0.25$, so $k = 2$ of the eight sampled encodings receive gradient together with the canonical one, three backward passes per prompt in place of nine. Variants that compare along a continuation use a reference continuation of up to $128$ tokens. All arms are fully fine-tuned for $700$ steps at learning rate $10^{-5}$ with gradient accumulation $8$. On OLMo-2-7B the layer sets for representation matching are $\mathcal{S} = \{32\}$, $\{8,16,24,32\}$ and all $32$ layers. Appendix~\ref{app.sec:details} lists the remaining hyperparameters.

\noindent\textbf{Baselines.} Canonical-only self-distillation is the CE objective without retokenized prompts. StochasTok \citep{sims2025stochastok} retokenizes the instruction and the answer at rate $0.3$ and trains on the retokenized answer tokens. Context-perturbed SFT (Section~\ref{sec:consistency}) trains on Alpaca answers and reference refusals over the same nine encodings, weighted uniformly. Appendix~\ref{app.sec:ablations} reports further exposure baselines.

\noindent\textbf{Attacks and judging.} AdvTok \citep{geh2025adversarial} is searched separately against each model on the $200$ held-out AdvBench prompts, formatted with the model's chat template, with no system prompt. Responses are decoded greedily ($T{=}0$) or sampled eight times at $T{=}1$ and judged by Llama-Guard-3-8B \citep{inan2023llama}. Appendix~\ref{app.sec:judges} repeats the main table under the HarmBench classifier. Transfer is tested with BEAST \citep{sadasivan2024fast} and Charmer \citep{abadrocamora2024charmer}, with settings in Appendix~\ref{app.sec:attacks}. Section~\ref{sec:semantics_diversity} also attacks $200$ WildJailbreak prompts \citep{jiang2024wildteaming}, which lie outside the training distribution.

\noindent\textbf{Capability and over-refusal.} Section~\ref{sec:semantics_diversity} reports GSM8K accuracy \citep{cobbe2021training} as the paired change from the untrained model. Appendix~\ref{app.subsec:rev-capability} reports MMLU, GSM8K and IFEval for all three models. Refusal of benign prompts is measured on $100$ held-out Alpaca instructions and the $250$ safe XSTest prompts \citep{rottger2024xstest} with a keyword-prefix judge (Appendix~\ref{app.sec:judges}).

\section{Robustness requires diversity over tokenizations}
\label{sec:robustness}

Training on deeply retokenized prompts lowers AdvTok success under every objective we test. We then isolate what the reduction depends on (exposure to retokenized prompts, their depth, and harmful prompts in the mixture) and report over-refusal and transfer to other attacks.

\subsection{Consistency training defends against adversarial tokenization attacks}\label{subsec:asr}

\begin{table}
\centering
\small
\setlength{\tabcolsep}{2.8pt}
\newcommand{\ci}[1]{{\scriptsize$\,\pm$#1}}
\caption{Training on retokenized prompts lowers AdvTok success on OLMo-2-7B under every objective. The consistency objectives keep refusal of safe prompts near the untrained level. AdvTok columns give attack success on $200$ held-out AdvBench prompts at greedy ($T{=}0$) and sampled ($T{=}1$) decoding, and the last column gives refusal on the $250$ safe XSTest prompts. Values are means over training seeds (three for Consistency, KL and repr-next, one otherwise) $\pm$ half the width of a $95\%$ bootstrap interval over prompts.}
\label{tab:attacks}
\begin{tabular}{@{}lccc@{}}
\toprule
 & \multicolumn{2}{c}{AdvTok} & XSTest-safe \\
\cmidrule(lr){2-3}
 & $T=0$ & $T=1$ & refusal \\
\midrule
    No fine-tuning & $0.400$\ci{.067} & $0.458$\ci{.052} & $0.065$\ci{.024} \\
    Canonical-only self-distillation & $0.325$\ci{.065} & $0.426$\ci{.046} & $0.102$\ci{.029} \\
    StochasTok, $\alpha{=}0.3$ & $0.205$\ci{.055} & $0.336$\ci{.050} & $0.108$\ci{.029} \\
\midrule
\multicolumn{4}{@{}l}{\emph{Trained on retokenized prompts}} \\
    Context-perturbed SFT & $0.020$\ci{.018} & $0.062$\ci{.026} & $0.235$\ci{.039} \\
    Consistency, CE & $0.030$\ci{.022} & $0.073$\ci{.025} & $0.085$\ci{.027} \\
    Consistency, KL & $0.020$\ci{.015} & $0.043$\ci{.018} & $0.068$\ci{.018} \\
    Consistency, repr-next & $0.038$\ci{.020} & $0.033$\ci{.016} & $0.064$\ci{.022} \\
\bottomrule
\end{tabular}
\end{table}

Every objective trained on retokenized prompts lowers AdvTok success by roughly an order of magnitude on OLMo-2-7B (Table~\ref{tab:attacks}). Appendix~\ref{app.sec:rebuttal-results} shows the reduction on Llama-3.2-3B and gemma-3-4b. We detect no AdvTok difference among the CE, KL and repr-next objectives. Attack success varies more across training seeds than across objectives, with greedy rates between $\val{0.010}$ and $\val{0.080}$ over the three repr-next seeds. Table~\ref{tab:depth-series} reports repr-next with fewer supervised layers. 

\noindent\textbf{Robustness requires exposure to retokenized prompts.} Canonical-only self-distillation is the CE objective without retokenized prompts. It leaves attack success close to the untrained level, $\val{0.295}$ above that of the CE model.

\noindent\textbf{Shallow retokenization does not protect against the attack.} StochasTok's encoding sampler inflates the token count by $\val{1.28}\times$ on average, while the MDD sampler we use gives $\val{3.62}\times$. Trained with the shallower sampler, the KL model has an attack success of $\val{0.165}$, eight times its value with the MDD sampler and close to that of full-sequence StochasTok at a similar depth (Appendix~\ref{app.sec:ablations}).

\noindent\textbf{Harmful prompts are needed in the data mixture.} A model trained with the KL objective on benign prompts alone has an attack success of $\val{0.315}$, only a small reduction from the untrained model (paired difference $\val{-0.085}$). 
Although the benign-only KL model represents retokenized harmful prompts more like their canonical encodings than the untrained model does, removing about three quarters of the final-layer residual distance (Table~\ref{tab:residual-distance}), its refusal probe is no more robust under attack (Table~\ref{tab:probe-transfer}). After fine-tuning on mixed data, that distance nearly vanishes for harmful prompts. On benign prompts the two trained models (i.e. the benign-only model and the mixed-data model) are indistinguishable.

\noindent\textbf{Context-perturbed SFT approximately matches consistency training on AdvTok but loses more benchmark accuracy.} The SFT model is trained on the same nine encodings with uniform weighting, with Alpaca answers as benign targets. Its attack success is comparable to that of the CE objective with uniform weighting, which differs only in using reference continuations as benign targets (paired difference $\val{-0.025}$). It loses more accuracy than any consistency model on all three benchmarks, and more than three times as much on IFEval (Table~\ref{tab:rev-capability}). This agrees with the findings of \citet{yang-etal-2024-self}, who attribute forgetting during fine-tuning to the gap between a task dataset and the model's own output distribution and reduce it by training on the model's own responses. The SFT model's refusal of safe prompts varies with the training seed (Appendix~\ref{app.sec:overrefusal}).

\noindent\textbf{On OLMo-2-7B, consistency training leaves refusal of safe prompts near the untrained level.} The rate varies by a factor of two to three across training seeds (Appendix~\ref{app.sec:overrefusal}). On Llama-3.2-3B every fine-tuned model refuses more safe XSTest prompts than the untrained model (Appendix~\ref{app.subsec:rev-capability}). A model trained on harmful prompts alone refuses indiscriminately, declining more than a third of benign instructions.

\noindent\textbf{Full-sequence retokenization also protects, but the model then produces non-canonical tokens.} At high $\alpha$, full-sequence StochasTok lowers AdvTok success to the level of consistency training (Appendix~\ref{app.sec:stochastok}). The model trained at $\alpha = 5.5$ emits non-canonical tokens at a rate of $\val{0.42}$, as Proposition~\ref{prop:target} predicts. Context-only training keeps the rate under $\val{0.001}$.

\noindent\textbf{Transfer to suffix-based attacks depends on the objective.} On OLMo-2-7B and gemma-3-4b, BEAST success is higher than the untrained model's for the KL and repr-cont models and lower for the repr-next model (Table~\ref{tab:rev-attacks}). On Llama-3.2-3B every trained model has a lower rate. The repr-next objective is therefore the only one that lowers both AdvTok and BEAST success on all three models. However, it also loses the most benchmark accuracy on the two smaller models (Appendix~\ref{app.subsec:rev-capability}). Results for Charmer attacks are mixed across models.

\noindent\textbf{Robustness transfers to character noise.} \citet{zheng2025broken} conjectured that models read non-canonical token sequences as misspelled text. If so, the invariance learned over retokenizations should extend to spelling noise, which also changes the token sequence of a prompt but, unlike retokenization, changes its bytes as well. We perturb the held-out harmful prompts with random character substitutions, encode them canonically and decode greedily. Eight substitutions raise the untrained model's unsafe rate from $\val{0.000}$ to $\val{0.195}$, so spelling noise alone is an effective attack. The KL and repr-next models stay at or below $\val{0.010}$, although their training never presented character edits. The transfer depends on the harmful prompts in the mixture: the model trained with the KL objective on benign prompts alone reaches $\val{0.265}$, above the untrained model. The trained models do not obtain this robustness by refusing more. Their refusal of perturbed benign instructions is unchanged, and that of perturbed safe XSTest prompts rises by $\val{0.064}$ (Appendix~\ref{app.sec:domains}).
 
\section{Generalization requires diversity over semantics}
\label{sec:semantics_diversity}

\revstart%
Prompt diversity drives most of the generalization of fine-tuned language models \citep{chung2024scaling, zhang2024instruction}, while adding responses to a fixed set of prompts shows sharply diminishing returns \citep{wang2022super0naturalinstructions0, zhou2023lima0, yu2023metamath0}. However, current approaches do not distinguish between surface token sequence and semantics diversity. However, we can imagine retokenizing the prompts so that they share no common tokens, while still corresponding to the same byte sequence. If prompts count toward data diversity by their surface form alone, retokenization could serve as a method to increase the dataset entropy at almost no cost. We study whether it holds when learning from finite data, using Llama-3.2-3B-Instruct, following the exact same setup.
\revend%

\revstart%
\noindent\textbf{Prompt diversity vs.\ retokenization budget.} To separate the effects, we vary both, the number of distinct prompts and the number of retokenizations for each prompt. We lower the prompt diversity to $f \in \{\nicefrac12, \nicefrac14, \nicefrac18\}$ by keeping a nested fraction $f$ of the harmful prompts per resulting dataset but shrink the dataset size by $f$, so the number of \emph{distinct} prompts, harmful and benign, shrinks by $f$. 

Then, we either sample additional re-tokenizations to compensate for the reduciton of the dataset size, so that the number of distinct \emph{token sequences} is equal to the full dataset, or keep the number of retokenizations the same, which reduces the number of distinct \emph{token sequences} by $f$ as well. If distinct token sequences could serve as substitutions for data diversity, we would expect downstream results in runs with compensated retokenizations to match full dataset training closely. If generalization requires diversity over distinct semantics, we would expect the results to resemble training on reduced datasets closer. Appendix~\ref{app.sec:prompt-diversity} provides the detailed description of the setup.

\revend%

\begin{figure}[t]
    \centering
    \includegraphics[width=\linewidth]{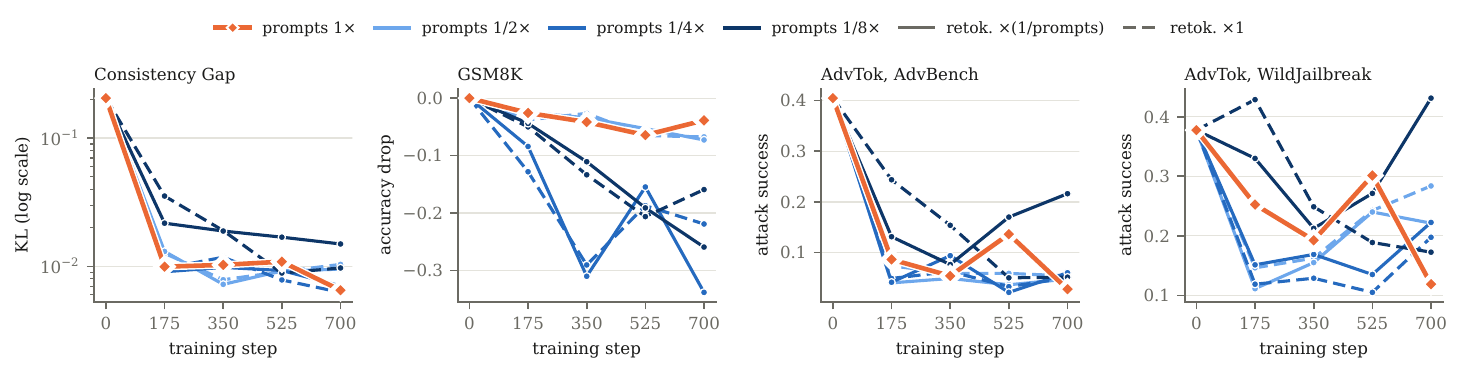}
    \caption{\rev{\textbf{Generalization requires diversity over prompts}. With fewer training prompts, the generalization against adversarial tokenization attacks for Llama-3.2-3B-Instruct trained with the KL consistency objective is worse and, at low prompt diversity, its capability loss increases, even when the number of \emph{distinct training samples} is kept intact. Solid lines are the runs with additional re-tokenizations in place of removed prompts, which keeps the same number of samples per dataset as the full run, while dashed lines  represent checkpoints trained on propotionally smaller sample sizes.}}
    \label{fig:prompt-diversity}
\end{figure}

\revstart%
\noindent\textbf{Results.} We find that retokenizations can not be used as substitutions for prompts (Figure~\ref{fig:prompt-diversity}). At the same number of training samples, every reduction in prompt diversity leaves the model more vulnerable than the full run, both on held-out prompts of the training distribution and on WildJailbreak. Nor does a larger retokenization budget on fewer prompts improve on leaving it unchanged. The matched arm follows the unmatched one while prompt diversity is high, and falls behind it once diversity shrinks, where it is both easier to attack and less accurate on GSM8K. Every arm nevertheless fits the objective, closing most of the consistency gap on held-out prompts (Appendix~\ref{app.sec:prompt-diversity}), so the arms differ in what they generalize to, not in how well they fit. Representation matching with continuation supervision (repr-cont) behaves differently: its robustness and accuracy barely depend on prompt diversity (Appendix~\ref{app.sec:prompt-diversity}).
\revend%

\revstart%
\noindent\textbf{Retokenization as augmentation.} In the terms of \citet{gontijo-lopes2020affinity}, an augmentation helps in-distribution generalization when it is diverse, meaning hard for the model to fit, and stays in affinity with the training distribution. Retokenization is neither: it is far out of distribution for the untrained model, yet \rev{consistency training fits it at every level of prompt diversity,} because it carries no content the model has not already fit. Such an augmentation cannot add data, as Section~\ref{sec:conditional} predicts. What it can do is enforce an invariance the model lacks \citep{geiping2022much, gao2023targeted}: \rev{every arm of Figure~\ref{fig:prompt-diversity} closes most of the consistency gap on held-out prompts, whatever its prompt diversity (Appendix~\ref{app.sec:prompt-diversity}).} For the output-based objective, generalization therefore needs diversity over the semantics of the prompt, which retokenization does not supply. Representation matching does not show this dependence in the range we test (Appendix~\ref{app.sec:prompt-diversity}).
\revend%

\section{Probing the safety and semantics}
\label{sec:probes}

\revstart%
Under adversarial tokenization refusal-probe accuracy falls below chance, while topic stays above chance. After consistency training the refusal signal persists under attack, in the same subspace.
\revend%

\noindent\textbf{Setup.} A \emph{probe} is a logistic classifier applied to the residual stream at the final prompt token, the last chat-template token after the assistant marker, where refusal rather than harmfulness is encoded \citep{arditi2024refusal,zhao2025harmfulness}. We fit two probes on the untrained model's activations for canonically tokenized prompts. The \emph{refusal probe} separates harmful from benign prompts. The \emph{topic probe} assigns each harmful prompt to one of six topics, and we call its accuracy \emph{topic decodability}; always guessing the largest topic gives $\val{0.35}$ on the held-out prompts. Both probes are fit once, on a training split of prompts, and then applied unchanged to every model (untrained or fine-tuned) and every to tokenization. When a frozen probe still works on a fine-tuned model, it indicates that the information lies along the same directions in activation space as it did before training. Attacked prompts use the tokenizations AdvTok found against the model being probed. Probes are read at layer 32 unless stated. Below, we report classification accuracies of the probes. Appendix~\ref{app.sec:probes} describes the protocol in detail.

\revstart%
\noindent\textbf{Suppression.} On the untrained model the refusal probe classifies every canonical harmful prompt correctly ($\val{1.000}$). On attacked tokenizations its accuracy is $\val{0.317}$, so it labels $\val{68}\%$ of them as benign. Part of this is an effect of depth. Random retokenizations matched to the attack's depth on each prompt already lower the probe to $\val{0.603}$. The attack lowers it further (difference $\val{-0.286}$). Behavior follows the same pattern: with no search, random retokenizations with 21 to 40 added tokens induce an unsafe response on $\val{0.228}$ of attempts, against $\val{0.400}$ for the search and $\val{0.000}$ for the canonical tokenization.
\revend%

\begin{wraptable}{r}{0.45\textwidth}
\centering
\small
\vspace{-1.2em}
\begin{tabular}{@{}lc@{}}
\toprule
 & \begin{tabular}[b]{@{}c@{}}Probe accuracy\\ under attack\end{tabular} \\
\midrule
No fine-tuning & $0.317$ \\
Consistency, KL, benign-only & $0.333$ \\
Consistency, CE (uniform) & $0.794$ \\
Consistency, KL & $0.921$ \\
Consistency, repr-next & $0.984$ \\
\bottomrule
\end{tabular}
\caption{\textbf{Consistency training on mixed data repairs brittleness of the refusal signal under retokenization attacks.} Accuracy at layer 32 on the 63 held-out harmful prompts, under the attack searched against each model. On the untrained model the probe reads $1.000$ on canonical tokenizations and $0.603$ on depth-matched random retokenizations.}
\label{tab:probe-transfer}
\end{wraptable}
\noindent\textbf{Adversarial search does not target topic decodability.} Topic decodability is $\val{0.762}$ on canonical prompts and $\val{0.587}$ under the attack, above the majority-class rate of $\val{0.35}$. Depth-matched random retokenization gives $\val{0.571}$, which the attack does not lower further (difference $\val{+0.016}$). Therefore the search, intended to target safety behavior, indeed selectively lowers refusal-probe accuracy but not topic decodability (beyond the effect of retokenization depth).

\noindent\textbf{Persistence after consistency training.} 
The frozen probe still classifies attacked harmful prompts as harmful on every model trained on mixed data, with accuracy between $\val{0.79}$ and $\val{0.98}$ (Table~\ref{tab:probe-transfer}). The probe was fit before any training, so the refusal signal in the trained models lies along the directions it occupied in the untrained model. Benign-only training does not produce this persistence.

\noindent\textbf{Causal steering.} Following \citet{arditi2024refusal}, we take the difference in mean activation between harmful and benign canonical prompts at layer 16 as a refusal direction and add it from the final prompt position onward, scaled to the mean projection of canonical harmful prompts. We apply it while the untrained model generates from the $80$ attacked tokenizations that produced an unsafe greedy response. The model then refuses on $\val{0.787}$ of them and the unsafe rate falls to $\val{0.050}$. Applying a random direction of the same norm induces no refusal behavior and leaves the unsafe compliance rate at $\val{0.925}$. The attack therefore leaves the downstream refusal mechanism intact but fails to activate it. 

\section{Conclusion}
\label{sec:conclusion}

\revstart%

Every byte string admits exponentially many valid tokenizations, and LLMs trained on the canonical one alone still acquire a partial invariance over the others. However, such invariance is partial, uneven, and adversarially fragile. We measured that fragility with the target NLL and traced it mechanistically to the suppression of the linearly decodable refusal signal. Unsupervised consistency training over retokenizations of the prompt keeps it in place under attack, in its original subspace, and on OLMo-2-7B lowers greedy AdvTok success from $0.400$ to $0.020$. Our experiments also show the limits of the method. Robustness requires deep retokenization and harmful prompts in the mixture. The robustness transfers to character noise, but only transfers to adversarial suffixes for all-layer representation matching at the last prompt position. Under the output-based objective, retokenized prompts do not substitute for new ones, while representation matching is insensitive to the number of prompts in the range we test.
\revend%

\newpage

\bibliography{tokenization}
\bibliographystyle{iclr2027_conference} 

\appendix
\clearpage

\section*{Limitations}

The probing and steering analyses were run on a single model, \texttt{OLMo-2-1124-7B-Instruct}; Appendix~\ref{app.sec:rebuttal-results} repeats only the attack and capability evaluations on Llama-3.2-3B-Instruct and gemma-3-4b-it. Three training seeds exist for the KL and repr-next arms and one for the others. Across those seeds greedy AdvTok success ranges from $0.010$ to $0.030$ for KL and from $0.010$ to $0.080$ for repr-next, so we do not rank the two on AdvTok. Attack success also depends on the judge (Appendix~\ref{app.sec:judges}). AdvTok requires control over token IDs, so it applies to open-weight models and to APIs that accept token IDs. However, an attacker who can submit token IDs does not need access to the model's output distribution, as random retokenization alone succeeds on $\val{0.228}$ of attempts against the untrained model. The prompt-diversity study of Section~\ref{sec:semantics_diversity} uses one model and one training seed per arm, as do the gemma-3-4b results.

\section*{AI use statement}

In this work, we used generative AI tools to assist with proving mathematical
claims, including checking and developing intermediate derivations and proof
steps. We did not use generative AI tools for other research ideation or
execution tasks, and the generation of synthetic datasets was not applicable
to this work. Additionally, we used generative AI tools for literature
retrieval and discovery, drafting parts of the manuscript, and improving the
clarity and readability of the writing.

All AI-assisted mathematical arguments and derivations were manually reviewed
and checked by the authors for correctness. References and related work
identified with the assistance of AI tools were verified against the original
sources, and all AI-assisted text was reviewed and edited by the authors. We
take responsibility for the final content of this work, including text, claims,
and artifacts produced with the aid of generative AI.

\section{Proofs for Section~\ref{sec:conditional}}
\label{app.sec:proofs}

\noindent\textbf{Setup.} Since $\kappa$ is deterministic, every tokenization decodes to a unique byte sequence, so $P_\theta(\boldx\mid\boldt)=1[\kappa(\boldt)=\boldx]$. For $P_\theta$ defined over complete token sequences, Bayes' theorem gives, for any $\boldt\in\tokenset(\boldx)$,
\begin{equation}
\label{eq:bayes_tokenization}
    P_\theta(\boldt\mid\boldx)=\frac{P_\theta(\boldt)}{P_\theta(\boldx)}\,1[\boldt\in\tokenset(\boldx)],
    \qquad P_\theta(\boldx)=\sum_{\boldt\in\tokenset(\boldx)}P_\theta(\boldt).
\end{equation}

\noindent\textbf{Proof of Proposition~\ref{prop:target}.} For any $\boldt\in\tokenset(\boldx)$ and any $q$ supported on $\tokenset(\boldx)$, Eq.~\eqref{eq:bayes_tokenization} yields the pointwise identity
\begin{equation}
\label{eq:pointwise_decomposition}
    -\log P_\theta(\boldt)
    = -\log P_\theta(\boldx) - \log q(\boldt\mid\boldx) + \log\frac{q(\boldt\mid\boldx)}{P_\theta(\boldt\mid\boldx)}.
\end{equation}
Taking the expectation over $T\sim q(\cdot\mid X)$, the second term is the entropy of the tokenizer and the third the divergence between the tokenizer and the model's conditional over retokenizations, which gives the equality in Eq.~\eqref{eq:entropy_bound}. The inequality follows from the non-negativity of the KL divergence, with equality if and only if $P_\theta(\cdot\mid\boldx)=q(\cdot\mid\boldx)$ for all $\boldx$. \hfill$\square$

The entropy term is at most $\log|\tokenset(\boldx)|$, attained when $q$ is uniform over retokenizations. The gap exists entirely at the token level: the string-level term $\mathbb{E}_X[-\log P_\theta(X)]$ is unaffected, and marginalizing over $\tokenset(\boldx)$ recovers the distribution over byte sequences, although $\tokenset(\boldx)$ grows exponentially in length and exact marginalization is impractical at scale \citep{cao-rimell-2021-evaluate, chirkova-etal-2023-marginalize}. Under canonical tokenization $H(q(\cdot\mid\boldx))=0$ for every $\boldx$, and the remaining gap $\mathbb{E}_X[-\log P_\theta(\boldt_c(X)\mid X)]$ vanishes as the model concentrates each string's mass on its canonical tokenization; this holds approximately for trained LLMs, which usually (but not always) assign small probability to non-canonical sequences \citep{geh-etal-2024-signal}, and exactly for models free of spurious ambiguity \citep{gastaldi2024foundations}. At the optimum of a stochastic target tokenization, $P_\theta(\cdot\mid\boldx)=q(\cdot\mid\boldx)$, the model remains a consistent estimator of the marginal over strings, but its token-level loss is floored at $\mathbb{E}_X[H(q(\cdot\mid X))]$ above the string-level loss, token-level likelihoods become poor estimates of string likelihoods without the marginalization of Eq.~\eqref{eq:bayes_tokenization}, and any sampling adjustment acting on per-token distributions, such as temperature or nucleus sampling, operates on a string-dependent entropy that carries no information about the bytes.

\noindent\textbf{Proof of Proposition~\ref{prop:prompt}.} During pretraining the model learns the unconditional $P_\theta(\boldt)$; during fine-tuning and post-training it is conditioned on a prompt. Consider prompt--continuation pairs $(Z,X)\sim\mu$ with two tokenizers, $(\tau_Z,\kappa)$ stochastic on the prompt and $(\tau_X,\kappa)$ canonical on the continuation, so $\boldt^Z\sim q(\cdot\mid Z)$ and $\boldt^X=\boldt_c(X)$, and the objective
\begin{equation}
\label{eq:prompt_retok_objective}
    \mathcal{L}(\theta)=\mathbb{E}_{(Z,X)\sim\mu,\;\boldt^Z\sim q(\cdot\mid Z)}\!\left[-\log P_\theta\big(\boldt_c(X)\mid\boldt^Z\big)\right].
\end{equation}
Since $\boldt^Z$ is drawn from $Z$ alone, $X\perp\boldt^Z\mid Z$, and because $\kappa$ is deterministic, $\mu(\cdot\mid\boldt^Z)=\mu(\cdot\mid\kappa(\boldt^Z))=\mu(\cdot\mid Z)$. Applying Eq.~\eqref{eq:pointwise_decomposition} to the continuation with the point-mass tokenizer at $\boldt_c(X)$ gives
\begin{equation}
    -\log P_\theta\big(\boldt_c(X)\mid\boldt^Z\big) = -\log P_\theta\big(X\mid\boldt^Z\big) - \log P_\theta\big(\boldt_c(X)\mid X,\boldt^Z\big),
\end{equation}
and taking expectations, with $\mathbb{E}[-\log P_\theta(X\mid\boldt^Z)] = H_\mu(X\mid Z) + \mathbb{E}_{Z,\boldt^Z}[\mathrm{KL}(\mu(\cdot\mid Z)\,\|\,P_\theta(\cdot\mid\boldt^Z))]$, yields Eq.~\eqref{eq:prompt_retok_decomposition}. Both added terms are non-negative; they vanish if and only if, for $q$-almost every $\boldt^Z$, $P_\theta(\cdot\mid\boldt^Z)=\mu(\cdot\mid\kappa(\boldt^Z))$ on strings and $P_\theta(\boldt_c(\boldx)\mid\boldx,\boldt^Z)=1$, so the infimum is $H_\mu(X\mid Z)$, as under canonical training. Whenever $q(\boldt_c(Z)\mid Z)>0$ the minimizer also coincides with the canonical-training optimum on canonical prompts, and the same holds for the worst-case variant that replaces the expectation over $\boldt^Z$ by a supremum over $\tokenset(Z)$, which directly targets adversarial retokenizations of the prompt. \hfill$\square$

\section{Prompt diversity and retokenization budget}\label{app.sec:prompt-diversity}
{\color{revblue}
\noindent\textbf{Design.} All arms apply the training recipe of Table~\ref{tab:attacks} to Llama-3.2-3B-Instruct: $700$ steps, learning rate $10^{-5}$, one prompt per micro-batch, $28.6\%$ harmful draws, a reference continuation of up to $128$ tokens sampled at $T{=}1$ for every example, and CVaR at $\rho = 0.25$ with the canonical term always kept. The $320$ harmful training prompts are shuffled once with a fixed seed, and prompt diversity $f$ keeps the first $\lceil 320 f\rceil$, so smaller prompt sets are nested in larger ones. Gradient accumulation is $8f$, so the number of prompts drawn shrinks with $f$ while each is reused as often as in the full run (about five times for harmful prompts, once for benign ones). Matched arms sample $8/f$ retokenizations per prompt ($16$, $32$, $64$) and backpropagate the worst $2/f$; unmatched arms sample $8$ and backpropagate $2$. Keeping the objective of Section~\ref{sec:consistency} unchanged has one side effect: the canonical term carries weight $1/(k+1)$, so in the matched arms it weighs $\nicefrac15$, $\nicefrac19$ and $\nicefrac1{17}$ instead of $\nicefrac13$. Both objectives share these weights.

\begin{table}[ht]
\centering
\small
\newcommand{\ci}[1]{{\scriptsize$\,\pm$#1}}
\caption{\textbf{The values behind Figures~\ref{fig:prompt-diversity} and~\ref{fig:prompt-diversity-repr}.} The columns ``retok.\ $\times 1/p$'' and ``retok.\ $\times 1$'' are the matched and unmatched arms, and $p$ is the prompt diversity $f$; attack success at $T{=}1$, GSM8K as the change from the untrained model, consistency gap on held-out harmful prompts. Untrained model: $0.394$ (AdvBench), $0.324$ (WildJailbreak), $0.759$ (GSM8K), $0.2049$ (gap). $\pm$: half-width of a $95\%$ interval.}
\label{tab:prompt-diversity}
\resizebox{\linewidth}{!}{%
\begin{tabular}{@{}llcccccccc@{}}
\toprule
& & \multicolumn{2}{c}{AdvTok, AdvBench} & \multicolumn{2}{c}{AdvTok, WildJailbreak} & \multicolumn{2}{c}{GSM8K, change} & \multicolumn{2}{c}{Consistency gap} \\
\cmidrule(lr){3-4}\cmidrule(lr){5-6}\cmidrule(lr){7-8}\cmidrule(lr){9-10}
Objective & Prompts ($p$) & retok.\ $\times 1/p$ & retok.\ $\times 1$ & retok.\ $\times 1/p$ & retok.\ $\times 1$ & retok.\ $\times 1/p$ & retok.\ $\times 1$ & retok.\ $\times 1/p$ & retok.\ $\times 1$ \\
\midrule
KL & $1\times$ & \multicolumn{2}{c}{$0.022$\ci{.010}} & \multicolumn{2}{c}{$0.161$\ci{.035}} & \multicolumn{2}{c}{$-0.075$\ci{.023}} & \multicolumn{2}{c}{$0.0058$} \\
 & $\nicefrac{1}{2}\times$ & $0.057$\ci{.018} & $0.059$\ci{.017} & $0.201$\ci{.035} & $0.253$\ci{.039} & $-0.073$\ci{.024} & $-0.067$\ci{.024} & $0.0095$ & $0.0104$ \\
 & $\nicefrac{1}{4}\times$ & $0.064$\ci{.020} & $0.054$\ci{.012} & $0.218$\ci{.033} & $0.206$\ci{.034} & $-0.338$\ci{.030} & $-0.219$\ci{.028} & $0.0070$ & $0.0063$ \\
 & $\nicefrac{1}{8}\times$ & $0.206$\ci{.035} & $0.060$\ci{.018} & $0.399$\ci{.044} & $0.178$\ci{.032} & $-0.259$\ci{.028} & $-0.159$\ci{.026} & $0.0149$ & $0.0097$ \\
\midrule
Repr., last position & $1\times$ & \multicolumn{2}{c}{$0.013$\ci{.008}} & \multicolumn{2}{c}{$0.095$\ci{.030}} & \multicolumn{2}{c}{$-0.024$\ci{.021}} & \multicolumn{2}{c}{$0.0132$} \\
+ answer & $\nicefrac{1}{2}\times$ & $0.016$\ci{.013} & $0.026$\ci{.018} & $0.086$\ci{.028} & $0.056$\ci{.023} & $-0.050$\ci{.023} & $-0.045$\ci{.023} & $0.0095$ & $0.0106$ \\
 & $\nicefrac{1}{4}\times$ & $0.015$\ci{.010} & $0.026$\ci{.017} & $0.028$\ci{.016} & $0.062$\ci{.024} & $-0.081$\ci{.023} & $-0.080$\ci{.023} & $0.0100$ & $0.0119$ \\
 & $\nicefrac{1}{8}\times$ & $0.013$\ci{.012} & $0.018$\ci{.014} & $0.056$\ci{.025} & $0.104$\ci{.032} & $-0.056$\ci{.023} & $-0.039$\ci{.023} & $0.0067$ & $0.0075$ \\
\bottomrule
\end{tabular}}
\end{table}

\noindent\textbf{Evaluation.} AdvTok uses the settings of Appendix~\ref{app.sec:attacks} with $8$ samples at $T{=}1$; WildJailbreak is a fixed sample of $200$ prompts from its vanilla harmful split. Responses judged degenerate (fewer than three words, under $20\%$ distinct words, or a long response compressing below $0.2$ of its length) are counted as safe. GSM8K is the full test set ($1{,}319$ questions), 8-shot chain of thought with flexible answer extraction and no chat template, scored as the per-question paired difference from the untrained model. The consistency gap is the mean KL divergence between a model's next-token distributions after the canonical prompt and after a retokenization of it, along a $32$-token greedy continuation of the canonical prompt, over $200$ held-out harmful and $200$ held-out benign prompts with four retokenizations each, the same for every arm.

\begin{figure}[htbp]
    \centering
    \includegraphics[width=\linewidth]{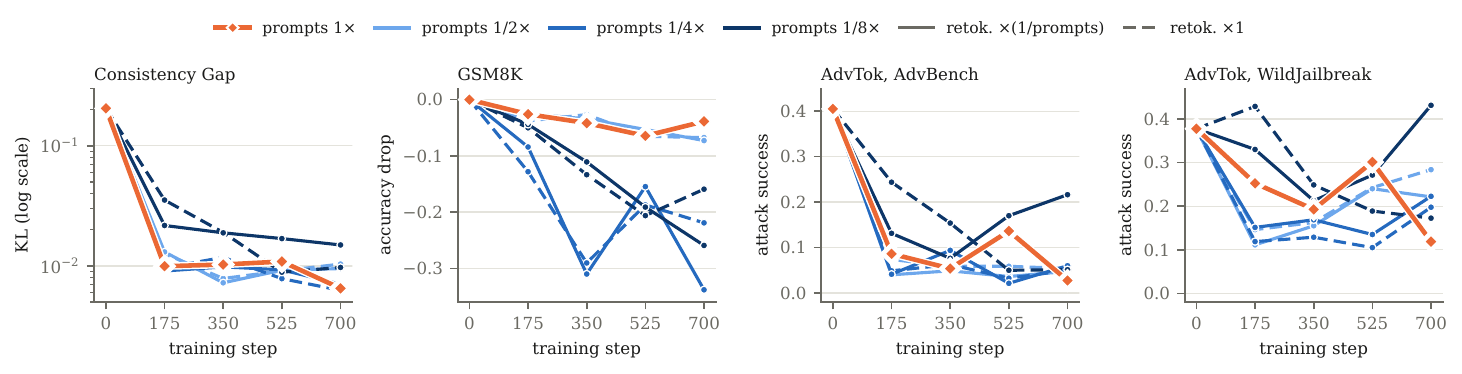}
    \caption{Figure~\ref{fig:prompt-diversity} (KL objective) redrawn on the vertical axes of Figure~\ref{fig:prompt-diversity-repr}, for comparison with representation matching.}
    \label{fig:prompt-diversity-kl-shared}
\end{figure}

\begin{figure}[htbp]
    \centering
    \includegraphics[width=\linewidth]{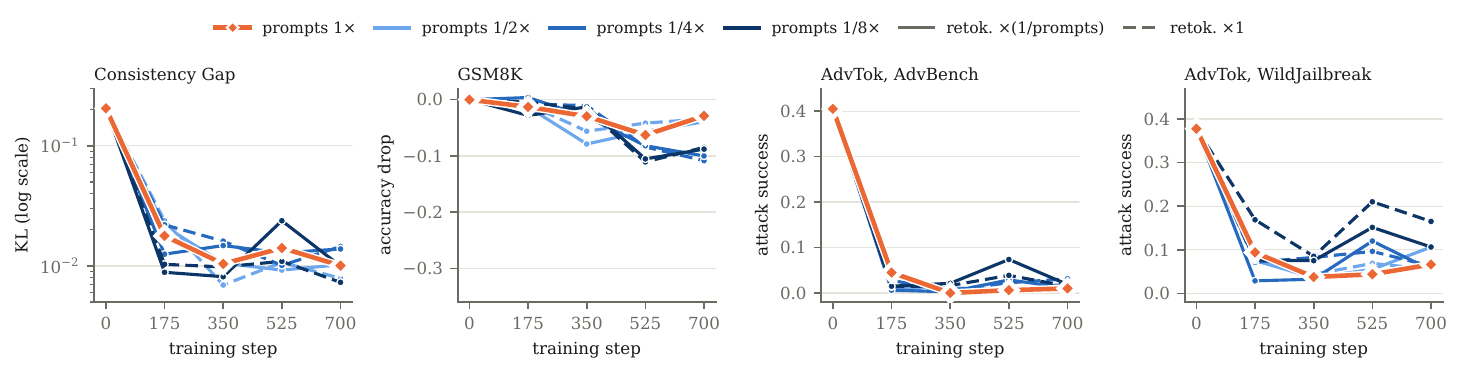}
    \caption{With representation matching, attack success and accuracy stay close to those of the full run at every level of prompt diversity and retokenization budget. Axes and line styles are as in Figure~\ref{fig:prompt-diversity}.}
    \label{fig:prompt-diversity-repr}
\end{figure}

\noindent\textbf{Representation matching is insensitive to prompt diversity.} The representation objective here supervises the residual stream at every layer, at the last prompt position and along the reference continuation. Its attack success on held-out AdvBench does not depend on prompt diversity or on the retokenization budget, no arm with fewer prompts is easier to attack on WildJailbreak than the full run, and GSM8K accuracy stays close to the full run's (Figure~\ref{fig:prompt-diversity-repr}, Table~\ref{tab:prompt-diversity}). The canonical-term weights are the same as for KL, so the weaker canonical anchor of the matched arms does not by itself explain the accuracy loss of the KL arms. One account, which we have not tested, follows from Appendix~\ref{app.sec:details}: to second order the KL objective is a squared error in the Fisher metric of the next-token law, which concentrates the penalty on residual directions the unembedding reads, the same directions that carry the model's behavior on other tasks; fitting few prompts harder therefore moves them, whereas the representation loss spreads its penalty over all directions, normalized per layer.
}

\section{Experimental Details and Hyperparameters}
\label{app.sec:details}

\subsection{Retokenization sensitivity sweep}

The twenty constructed sentences are matched in length and syntactic frame,
and every valid split position of the canonical tokenization of each is swept,
giving $526$ positions. Corpus replication draws the first 100 rows of
WikiText and of a Python code corpus and applies the same sweep. Cascade
paths and the corpus statistics are restricted to positions whose per-token
probability under the base model falls between $\val{0.05}$ and $\val{0.95}$, which
excludes positions whose target is near-deterministic or near-impossible; the
constructed statistics reported in Section~\ref{subsec:perturbation} are
unfiltered, and applying the band there leaves $415$ positions across sixteen
sentences. All measurements are teacher-forced in float32.

\subsection{Consistency training}

Only the instruction span is retokenized, and the chat-template tokens stay canonical. For each prompt we sample $8$ non-canonical tokenizations of that span uniformly from the MDD, which with the canonical encoding gives $9$ encodings. The reference continuation is sampled once at the canonical encoding, at temperature $1$ with no top-$k$ or nucleus truncation and at most $128$ new tokens, and is held fixed across the nine encodings. CVaR at $\rho = 0.25$ ranks the $8$ sampled encodings by the arm's own loss without gradients and backpropagates the worst $2$ together with the canonical encoding. The uniform-selection ablation of Appendix~\ref{app.sec:cvar} sets $\rho = 1$ and backpropagates all nine.

All full fine-tuning arms train for $700$ steps with one prompt per micro-batch and gradient accumulation $8$, so a run draws $5{,}600$ prompts. The learning rate is $10^{-5}$ with $20$ warmup steps, the gradient norm is clipped at $1.0$, and the optimizer is 8-bit AdamW. The benign-only and harmful-only results quoted in the main text also use $700$ steps; Table~\ref{tab:overrefusal} reports earlier $500$- and $200$-step runs of those two arms.

Each draw is harmful with probability $0.286$, which in the main run gave $1{,}547$ harmful and $4{,}053$ benign draws. The harmful pool is the $320$ AdvBench prompts outside the $200$ held out, each drawn about five times. The benign pool is $30{,}861$ Alpaca prompts, of which $3{,}793$ are drawn. Retokenizations were sampled from an unseeded random generator, so a rerun with the same training seed reproduces the prompt order but neither the sampled tokenizations nor the reference continuations.

\noindent\textbf{Baselines.} Context-perturbed SFT (Section~\ref{sec:consistency}) trains on Alpaca answers and reference refusals over the same nine encodings, weighted uniformly. The LoRA arms use rank $16$ and scaling factor $32$ on query, key, and value projections, at learning rate $2.5 \times 10^{-6}$.

\noindent\textbf{Matching distributions and representations.}
Since the expectation of the score function vanishes \citep{martens2020natural}, to second order in $\delta = h^{\ell_{\max}}_{\theta}(\boldt_i)-h^{\ell_{\max}}_{\mathrm{ref}}(\boldt_c)$ where $h^{\ell_{\max}}$ is the final layer hidden state, the forward KL equals $\tfrac12\,\delta^{\top}G\,\delta$ with $G=W^{\top}(\mathrm{diag}(p)-pp^{\top})W$ the Fisher information of the next-token law in residual coordinates \citep{hsu2022closer,amari2019fisher}. The output-based consistency objective is thus a squared error in the metric defined by $G$, while $D_{\mathrm{rep}}$ uses the identity matrix as a metric. Since $\delta^{\top} G  \delta=\operatorname{Var}_{y\sim p}[(W \delta)_y]$, $G$ is zero on $\ker W$ and is small along logits of tokens that are assigned low probability by the model, so $D_{\mathrm{rep}}$ also penalizes residual directions that the un-embedding head $W$ is invariant to. Although this might be valuable from the perspective of inner (belief) consistency, \citet{hsu2022closer} found that in distillation scenarios behavioral matching leads to a lower approximation error and better generalization w.r.t.\ the reference model than Fisher-weighted and representation matching. It is therefore somewhat unexpected that single-position representation matching is the only objective that lowers BEAST success on all three models (Table~\ref{tab:rev-attacks}).

\section{Splittability Analysis}
\label{app.sec:splittability}

\begin{figure}
    \centering
    \includegraphics[width=0.85\linewidth]{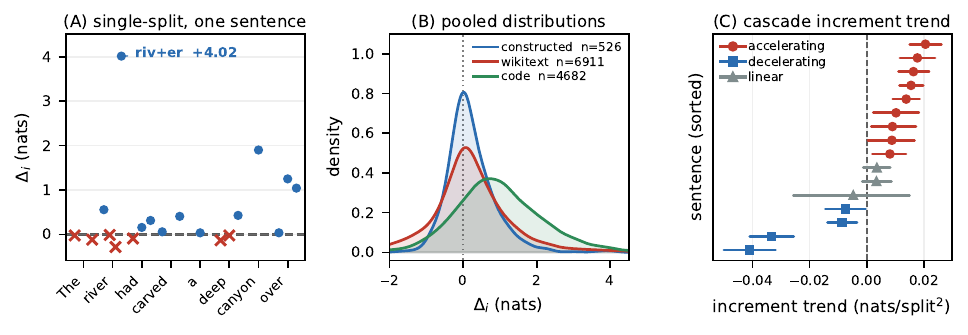}
    \caption{Retokenization sensitivity under OLMo-2-1124-7B base. \textbf{(A)} Single-split sensitivity $\Delta_i$ by position, for one test sentence (``The river had carved a deep canyon over"). \textbf{(B)} Pooled $\Delta_i$ distributions for constructed sentences vs.\ WikiText and code corpus-text replications (legend gives $n$ positions per domain; not normalized). \textbf{(C)} Trend in the per-split increment of $\mu(d)$ for the 16
    cascade-analyzed sentences (OLS slope of the increment against depth,
    nats/split$^2$; 95\% bootstrap intervals over paths). Positive values
    accelerate, negative decelerate; intervals containing zero are
    indistinguishable from linear.}
    \label{fig:sensitivity}
\end{figure}

All measurements in this appendix use the base model \texttt{allenai/OLMo-2-1124-7B} \citep{olmo2025two}, teacher-forced in float32. Sweep settings are in Appendix~\ref{app.sec:details}, and Section~\ref{app.subsec:posttraining} repeats the sweep on post-trained checkpoints.

\subsection{Single splits}
\label{app.subsec:residual_sensitivity}

A token can admit several two-piece decompositions. For a token sequence $\boldt = (t_1, \dots, t_k)$, a \emph{split position} is a pair $(i, c)$ such that $t_i = uv$ at cut point $c$ with $u, v \in V$. Applying the split yields
\begin{equation}
\boldt^{(i,c)} = (t_1, \dots, t_{i-1},\, u,\, v,\, t_{i+1}, \dots, t_k)
\in \tokenset(\boldx),
\end{equation}
and we write
\begin{equation}
\Delta_{i,c}(\boldt) := \fread\big(\boldt^{(i,c)}\big) - \fread(\boldt).
\end{equation}
We measure splits of the canonical tokenization only, and Section~\ref{subsec:perturbation} writes $\Delta_i$ for $\Delta_{i,c}(\boldt_c)$. The sweep covers every split position of $\boldt_c$ in twenty constructed sentences ($526$ positions). The target NLL is additive over target positions and deterministic under teacher forcing, so per-split changes and their accumulation are well defined.

\noindent\textbf{Largest effect.} The largest single value occurs in ``The river had carved a deep canyon over,'' where splitting \texttt{river} into \texttt{riv} and \texttt{er} raises the target NLL by $\val{4.02}$ nats (Figure~\ref{fig:sensitivity}A), a $5.8\sigma$ residual under the signed regression below.

\noindent\textbf{Feature regression.} We regress the single-split sensitivity
on seven features of the split position, together with sentence fixed effects,
by ordinary least squares with standard errors clustered by sentence; features
and outcome are standardized over the $526$ positions, so coefficients read in
standard deviations. The features are the Zipf log-frequency of the canonical
piece being split, the lower and the higher Zipf frequency of its two halves,
the cut offset normalized by piece length, the byte length of the piece, an
indicator that one half is the bare space marker, and an indicator that the
piece belongs to a word already split into more than one piece. We fit the
signed sensitivity and its magnitude separately. Sentence identity alone
explains $\val{0.179}$ of the variance in the signed sensitivity and $\val{0.167}$ in its
magnitude; the seven features raise these to $\val{0.210}$ and $\val{0.232}$. For the
magnitude, the cut position, the rarity of the canonical piece and its byte
length lead on drop-one explained variance at $\val{0.025}$, $\val{0.021}$ and $\val{0.013}$,
comparable to one another and small in absolute terms, while every share
collapses for the signed outcome.

\noindent\textbf{Corpus replication.} WikiText prose closely matches the
constructed set under the same per-token probability band (median $\val{+0.13}$
against $\val{+0.12}$, negative fraction $\val{0.37}$ against $\val{0.376}$, Gini $\val{0.52}$ against
$\val{0.496}$ on $415$ constructed positions), while code shifts toward higher median
sensitivity ($\val{+0.29}$), fewer negative positions ($\val{0.31}$), and heavier
concentration (Figure~\ref{fig:sensitivity}B). The statistics quoted in
Section~\ref{subsec:perturbation} are unfiltered.

\subsection{Cascades}
\label{app.subsec:cascades}

We sample $50$ cascade paths to depth $20$ for each of the sixteen sentences within the per-token probability band of Appendix~\ref{app.sec:details}. Write $g$ for the change in target NLL from the canonical tokenization, $\mu_s(d)$ for its mean over the paths of sentence $s$ at depth $d$, and $\mu(d)$ for the mean of $\mu_s(d)$ over sentences. Over depths $1$ to $20$ the variance of $g$ splits exactly into the variance of $\mu(d)$ ($36\%$), of $\mu_s(d) - \mu(d)$ ($43\%$) and of $g - \mu_s(d)$ ($21\%$). On the raw target NLL, which also carries each sentence's canonical value, the shares are $13\%$, $79\%$ and $8\%$.

The spread between sentences grows faster with depth than the spread between cascades of one sentence. At depth $5$ the standard deviation of $\mu_s$ across sentences is $1.4$ nats, against a mean within-sentence standard deviation of $1.2$ nats across cascades. At depth $20$ the two are $3.9$ and $2.0$ nats. The share of variance between sentences at a fixed depth is $0.51$ at depth $5$, $0.66$ at depth $10$ and $0.75$ at depth $20$, with intervals from resampling sentences. Within a single sentence, depth accounts for $71\%$ of the variance in target NLL, path identity for $12\%$ and their interaction for $17\%$.

We fit a linear trend to the per-split increments $\mu_s(d) - \mu_s(d-1)$ by ordinary least squares against depth, with $95\%$ bootstrap intervals over paths. Nine sentences have a positive trend, so their splits have compounding effects. Four have a negative trend and three an interval containing zero (Figure~\ref{fig:sensitivity}C). Every $\mu_s(d)$ rises with depth, so a negative trend means diminishing effects per split while the NLL still rises. Two of the nine positive intervals have lower ends within $0.002$ nats per split squared of zero.

\subsection{Supervised Fine-tuning Reorganizes Sensitivity}\label{app.subsec:posttraining}

Partial invariance of LLMs to retokenized input emerges during pretraining \citep{jain2026emergent}, but the format-enforced distinction between prompt and response provided by supervised fine-tuning (SFT) is necessary for the model to respond fluently to non-canonical prompts \citep{zheng2025broken}. Repeating the sensitivity experiments above on \texttt{OLMo-2-1124-7B-Instruct} \citep{olmo2025two}, Olmo-3 \citep{olmo2025olmo3}, and Tulu-3-8B \citep{lambert2024tulu}, post-training reorganizes which positions are sensitive, and this is concentrated at base-to-SFT in all three families, with later stages largely inheriting the result. How much reorganization occurs varies by family, from about 10 percent of the base profile retained in Olmo-3 to 30 to 44 percent in OLMo-2 and Tulu-3. The distribution of sensitivity has no common pattern: it is unchanged in OLMo-2, shifts upward in magnitude in Olmo-3, and shifts downward in Tulu-3, where the negative fraction rises from 0.27 to 0.47 at the DPO \citep{rafailov2023direct} stage. Only the concentration of the positive tail is stable everywhere, with Gini between 0.48 and 0.51 at every checkpoint of every family.

Doing fine-tuning via SFT on aligned models can degrade alignment due to catastrophic forgetting \citep{mccloskey1989catastrophic, luo2023empirical}. Retokenized alignment replicates this phenomenon: SFT training the Instruct model on the Alpaca dataset \citep{taori2023alpaca} increased the AdvTok attack success from $\val{0.400}$ to $\val{0.500}$ at greedy decoding and from $\val{0.458}$ to $\val{0.537}$ at $t{=}1$ (Table~\ref{tab:baselines-full}). Retokenizing the instructions during the same training keeps it near the original level ($\val{0.410}$ at $1.30\times$ depth, $\val{0.330}$ at $3.10\times$).

\section{Cross-domain sensitivity}
\label{app.sec:domains}

Consistency training perturbs one family of inputs: MDD tokenizations of
instruction-domain prompts. We measure how far the resulting invariance extends
beyond it, to other text domains and to perturbations that are not
retokenizations.

\noindent\textbf{Character noise.} \citet{zheng2025broken} conjectured that models read non-canonical sequences as mis-spelled text, which would give retokenized and mis-spelled inputs shared representations. We perturb the 200 held-out harmful prompts with random character substitutions, encode the result canonically and decode greedily. Eight substitutions raise the untrained model's unsafe rate from $\val{0.000}$ to $\val{0.195}$. Output KL stays at $\val{0.000}$ (paired difference from the untrained model $\val{-0.195}$) and repr-next reaches $\val{0.010}$ ($\val{-0.185}$). Training on retokenizations therefore transfers to character edits, which the training never presented. Benign-only training does the opposite. Its unsafe rate under eight substitutions is $\val{0.265}$, above the untrained model's ($\val{+0.070}$). Responses are judged by Llama-Guard-3-8B.

\noindent\textbf{Text domains.} We repeat the
sensitivity sweep of Appendix~\ref{app.sec:splittability} on four
held-out prompt sets ($200$ prompts, roughly $3{,}000$ retokenized
variants each) and report the fraction of the untrained model's median
sensitivity that each arm removes (a dimensionless fraction; the
medians themselves are in nats). Output KL removes $\val{0.73}$ on
held-out instructions from the training domain and $\val{0.34}$--$\val{0.41}$ on
WikiText prose, code, and grade-school arithmetic; the reduction roughly
halves off domain and is uniform across the three. Canonical supervised
fine-tuning removes $\val{0.49}$ in domain while leaving code no better than no
training at all ($\val{3.06}$ against $\val{2.95}$ nats, intervals overlapping), so
the instrument registers domain-specific damage where it exists. Under
full character-level fragmentation the untrained model loses between
$\val{7.5}$ nats on instructions and $55$ on WikiText, and output KL halves
the loss in every domain.

\noindent\textbf{Benign prompts under noise.} Refusal on perturbed
benign prompts is unchanged: at eight substitutions output KL sits
at $\val{+0.000}$ against the untrained model on held-out
instructions ($n = 200$) and $\val{+0.064}$ on safe XSTest prompts
($n = 250$). Answer overlap degrades identically in every arm, trained or not, losing
$\val{0.20}$--$\val{0.25}$ of its unigram F1 under perturbation.

\section{CVaR}\label{app.sec:cvar}

Setting $\rho = 1$ removes the selection and backpropagates a uniform mean over all nine encodings. For output KL, the uniform run reaches $\val{0.020}$ at greedy decoding and $\val{0.036}$ under sampling. With the selection, three training seeds give $\val{0.010}$, $\val{0.030}$ and $\val{0.020}$ at greedy decoding. For cross-entropy, the uniform run reaches $\val{0.045}$ and $\val{0.072}$, and the run with the selection $\val{0.030}$ and $\val{0.073}$. These differences are within the variation across training seeds, so with eight sampled encodings the selection has no detectable effect on robustness. Its practical effect is on computation, since it backpropagates three of the nine encodings per prompt. Only $\rho \in \{0.25, 1\}$ was trained.

\section{Distributionally robust losses}\label{app.sec:dro}

The top-$k$ selection of Section~\ref{sec:consistency} is one member of a family of distributionally robust losses over the $n = N{-}1$ sampled encodings of a prompt. Each member replaces the uniform mean of the terms $d_1,\dots,d_n$ by a weighted mean $\sum_i w_i d_i$, with weights chosen adversarially within an uncertainty set around the uniform distribution. We compare four members, each applied within a prompt, with the canonical term $d_0$ kept outside the weighting at weight $1/(k+1)$ as in the main text.

\noindent\textbf{Top-$k$ CVaR.} The weights are $1/k$ on the $k$ largest terms and zero elsewhere. This plug-in estimate of $\operatorname{CVaR}_\rho$ is biased for finite $n$, because the same samples set the tail threshold and supply the gradient, and it understates the population CVaR.

\noindent\textbf{Rockafellar--Uryasev CVaR.} The variational form \citep{rockafellar2000cvar}
\begin{equation}
\operatorname{CVaR}_\rho(d) \;=\; \min_{\eta}\; \eta + \frac{1}{\rho\, n}\sum_{i=1}^{n} \big(d_i - \eta\big)_+
\end{equation}
gives an unbiased gradient for fixed $\eta$. We keep a single threshold $\eta$ across training and update it by its analytic gradient $1 - a/\rho$, where $a$ is the fraction of active encodings, so $\eta$ tracks the $(1-\rho)$ quantile. The terms are centered on their per-prompt mean before thresholding. Without centering, $\eta$ thresholds the pooled distribution over prompts and encodings, which then selects whole prompts rather than the worst encodings of each prompt. The number of encodings that receive gradient varies from step to step and equals $k$ on average.

\noindent\textbf{Entropic (softmax).} The weights are $w_i \propto \exp(d_i/\tau)$, the worst case over a KL ball around the empirical distribution, equivalently the entropic risk $\tau \log \frac1n \sum_i \exp(d_i/\tau)$ \citep{li2021tilted}. The limit $\tau \to 0$ gives the maximum and $\tau \to \infty$ the mean.

\noindent\textbf{$\chi^2$ ball.} The weights are $w_i \propto (d_i - \lambda)_+$, the worst case over a $\chi^2$ ball around the empirical distribution \citep{namkoong2016stochastic}. Unlike the softmax, it sets the weight of encodings below the threshold to exactly zero.

\noindent\textbf{Matched concentration.} The four losses differ in how strongly they concentrate on the tail as well as in the shape of the weighting. To compare the shapes alone, we set $\tau$ and $\lambda$ per prompt by bisection so that the weights sit at the divergence from uniform that top-$k$ produces, $\chi^2(w \,\|\, u) = n\sum_i (w_i - 1/n)^2 = 3$ at $n = 8$ and $\rho = 0.25$. The softmax and $\chi^2$ losses give gradient to all $n$ encodings and therefore backpropagate nine encodings per prompt instead of three.

\noindent\textbf{Results.} All arms use the KL objective and the MDD sampler with the remaining recipe of Table~\ref{tab:attacks} on Llama-3.2-3B-Instruct, and are evaluated as the Llama-3.2-3B KL model of Table~\ref{tab:rev-attacks}: AdvTok on the $200$ held-out AdvBench prompts, and BEAST and Charmer on the same prompts, AdvTok, BEAST and Charmer on $50$ WildJailbreak prompts, and the jailbreak wrappers and user-turn prefill of Appendix~\ref{app.subsec:rev-wrappers}, each with a retokenization search, which training did not use. The top-$k$ column is that model. Each loss is one training seed. All four losses bring AdvTok success on AdvBench close to zero. On the other attacks and datasets, the top-$k$ loss is never the best of the four, Rockafellar--Uryasev CVaR has the lowest greedy attack success in five of eight rows, and the $\chi^2$ ball loses the least benchmark accuracy.

\begin{table}[ht]
\centering
\caption{\textbf{The choice of robust loss affects transfer and capability, and top-$k$ CVaR is never the best of the four.} KL consistency training on Llama-3.2-3B-Instruct under four distributionally robust losses at matched concentration. Attack cells give attack success at $T{=}0$ / $T{=}1$ with degenerate responses excluded; XSTest-safe refusal is the keyword refusal rate on canonical prompts; capability is accuracy with the chat template. Bold marks the best of the four losses in each row. Backward passes are per prompt, the canonical encoding included.}
\label{tab:dro}
\begin{tabular}{@{}lcccc@{}}
\toprule
 & Top-$k$ CVaR & RU CVaR & Entropic & $\chi^2$ ball \\
\midrule
Backward passes per prompt & 3 & 3 (mean) & 9 & 9 \\
\midrule
\multicolumn{5}{@{}l}{\textit{Adversarial tokenization (training distribution)}} \\
\quad AdvTok, AdvBench & 0.005 / 0.022 & \textbf{0.000} / \textbf{0.018} & \textbf{0.000} / 0.021 & \textbf{0.000} / 0.028 \\
\midrule
\multicolumn{5}{@{}l}{\textit{Attacks and datasets not used in training}} \\
\quad BEAST, AdvBench & 0.225 / 0.410 & 0.235 / 0.407 & \textbf{0.155} / \textbf{0.387} & 0.220 / 0.473 \\
\quad Charmer, AdvBench & 0.170 / 0.328 & \textbf{0.030} / \textbf{0.193} & 0.080 / 0.203 & 0.185 / 0.255 \\
\quad AdvTok, WildJailbreak & 0.100 / 0.228 & \textbf{0.040} / \textbf{0.102} & 0.120 / 0.107 & 0.060 / 0.122 \\
\quad BEAST, WildJailbreak & 0.260 / 0.425 & 0.180 / 0.380 & \textbf{0.140} / \textbf{0.360} & 0.260 / 0.372 \\
\quad Charmer, WildJailbreak & 0.200 / 0.265 & 0.160 / \textbf{0.240} & \textbf{0.140} / 0.275 & 0.160 / 0.250 \\
\quad Wrapper + retok.\ search & 0.050 / 0.102 & \textbf{0.010} / \textbf{0.035} & 0.030 / 0.046 & 0.020 / 0.045 \\
\quad Prefill (user turn) + retok.\ search & 0.020 / 0.030 & \textbf{0.000} / 0.015 & \textbf{0.000} / 0.018 & \textbf{0.000} / \textbf{0.013} \\
\midrule
\multicolumn{5}{@{}l}{\textit{Over-refusal and capability}} \\
\quad XSTest-safe refusal & 0.190 / 0.163 & 0.140 / 0.211 & 0.100 / 0.168 & \textbf{0.080} / \textbf{0.136} \\
\quad MMLU (5-shot) & 0.551 & 0.534 & 0.539 & \textbf{0.564} \\
\quad GSM8K (8-shot CoT) & 0.684 & 0.697 & 0.677 & \textbf{0.702} \\
\quad IFEval (0-shot) & 0.477 & 0.495 & 0.530 & \textbf{0.553} \\
\bottomrule
\end{tabular}
\end{table}

\section{Exposure baselines}
\label{app.sec:ablations}

Consistency training is one of several ways to expose a model to
non-canonical tokenizations during fine-tuning. This appendix reports
the alternatives we trained, all on \texttt{OLMo-2-1124-7B-Instruct}
and all evaluated with a fresh AdvTok search against each model over the same $200$ held-out AdvBench prompts, under the protocol of Table~\ref{tab:attacks}, with responses judged by Llama-Guard-3-8B.

The arms differ along three axes that the table groups by, because
conflating them is misleading. The first is the \emph{training
mixture}: benign prompts alone, or the benign/harmful mixture used by our objective (harmful share $0.286$). The second is the \emph{supervision target}:
the dataset's own answers, or continuations generated by the frozen
reference. The third is \emph{exposure depth}, the mean inflation in
token count relative to canonical, which varies by an order of
magnitude across samplers.

\begin{table}[h]
\centering
\small
\begin{tabular}{llcc}
\toprule
 & depth & ASR greedy & ASR $t{=}1$ \\
\midrule
No fine-tuning & --- & 0.400 & 0.458 \\
\midrule
\multicolumn{4}{l}{\emph{Benign prompts, dataset answers}} \\
\quad Canonical SFT & $1.00\times$ & 0.500 & 0.537 \\
\quad Prompt-retokenized SFT & $1.30\times$ & 0.410 & 0.464 \\
\quad Prompt-retokenized SFT & $3.10\times$ & 0.330 & 0.387 \\
\midrule
\multicolumn{4}{l}{\emph{Benign prompts, reference continuations}} \\
\quad Consistency, output KL & $3.62\times$ & 0.315 & 0.402 \\
\midrule
\multicolumn{4}{l}{\emph{Mixed prompts, reference continuations}} \\
\quad Multi-tokenization LoRA, identical copies & $1.00\times$ & 0.490 & 0.577 \\
\quad Multi-tokenization LoRA, length-ratio buckets & $2.24\times$ & 0.200 & 0.223 \\
\quad Multi-tokenization LoRA, uniform lattice & $3.19\times$ & 0.120 & 0.177 \\
\quad StochasTok, $\alpha = 0.3$ & $1.28\times$ & 0.205 & 0.336 \\
\quad Consistency, output KL & $1.28\times$ & 0.165 & 0.237 \\
\quad Consistency, output KL & $3.62\times$ & 0.020 & 0.043 \\
\quad Consistency, repr-next & $3.62\times$ & 0.038 & 0.033 \\
\bottomrule
\end{tabular}
\caption{Exposure baselines. Depth is mean token-count inflation
relative to the canonical tokenization. The LoRA arms follow the
configuration of \citet{ghosh2026language} (rank $16$ on
query, key and value projections, learning rate $2.5\times10^{-6}$,
$30$ copies per example) applied to our training mixture rather than
to their task data, so they test their sampler within our setup and
are not a reproduction of their experiment. The adapters retokenize
the prompt span only; the treatment of the answer span in
\citet{ghosh2026language} is unspecified and is immaterial for their
single-token answers. The last two rows are means over three training seeds, as in Table~\ref{tab:attacks}; the other rows use one seed. Training budgets differ across groups ($700$ optimizer steps for the consistency and StochasTok rows, $3$ epochs for the SFT rows, $5$ epochs for the LoRA rows), so the rows are separate baselines and are not budget-matched.}
\label{tab:baselines-full}
\end{table}

Four observations follow.

\noindent\textbf{LoRA arms.} The control for the LoRA arms sees $30$ identical canonical copies of each example and reaches $\val{0.490}$ at greedy decoding. The two retokenized LoRA arms reach $\val{0.200}$ and $\val{0.120}$ (paired differences from the control $\val{-0.290}$ and $\val{-0.370}$). The mechanism reported by \citet{ghosh2026language} therefore carries over to an adversarial evaluation, which their work does not perform. The uniform-lattice sampler does better than the length-ratio buckets, the reverse of their ordering on task accuracy. It also retokenizes more deeply ($\val{3.19}\times$ against $\val{2.24}\times$).

\noindent\textbf{Fine-tuning without retokenization.} Two arms that train on dataset answers with canonical prompts end above the untrained model: full fine-tuning on Alpaca answers ($\val{0.500}$ against $\val{0.400}$ at greedy decoding) and the LoRA control ($\val{0.490}$). Canonical-only self-distillation, which trains on the reference model's own continuations, does not ($\val{0.325}$; Table~\ref{tab:attacks}).

\noindent\textbf{Depth.} Retokenizing prompts inside ordinary SFT on benign data leaves attack success unchanged at $\val{1.30}\times$ depth ($\val{0.410}$) and lowers it slightly at $\val{3.10}\times$ ($\val{0.330}$). Within the KL objective on the mixed data, depth has a large effect: $\val{0.165}$ at $\val{1.28}\times$ and $\val{0.020}$ at $\val{3.62}\times$. At the shallow depth the KL objective is close to full-sequence StochasTok ($\val{0.205}$).

\noindent\textbf{Harmful prompts.} At $\val{3.62}\times$ depth, the KL objective trained on benign prompts alone reaches $\val{0.315}$, against $\val{0.020}$ with harmful prompts in the mixture. The objective uses no refusal labels, and the reference model already refuses harmful prompts canonically. What the harmful prompts supply is the opportunity to preserve that behavior under retokenization.

\begin{table}[h]
\centering
\small
\newcommand{\cix}[1]{{\scriptsize$\,\pm$#1}}
\begin{tabular}{@{}lcc@{}}
\toprule
 & \multicolumn{2}{c}{AdvTok} \\
\cmidrule(l){2-3}
supervised layers & $T=0$ & $T=1$ \\
\midrule
$\{32\}$ & $0.050$\cix{.033} & $0.041$\cix{.023} \\
$\{8,16,24,32\}$ & $0.035$\cix{.025} & $0.046$\cix{.023} \\
all $32$ & $0.010$\cix{.013} & $0.009$\cix{.011} \\
\midrule
No fine-tuning & $0.400$\cix{.067} & $0.458$\cix{.052} \\
\bottomrule
\end{tabular}
\caption{Single-position representation matching as the supervised layer set
grows, on the $200$ held-out AdvBench prompts under the settings of
Appendix~\ref{app.sec:attacks}. Each row is one training seed.}
\label{tab:depth-series}
\end{table}

\begin{table}[h]
\centering
\small
\begin{tabular}{llcccc}
\toprule
Prompts, encodings & Arm & L8 & L16 & L24 & L32 \\
\midrule
Harmful, random & No fine-tuning & $0.413$ & $0.664$ & $0.546$ & $0.454$ \\
 & KL, benign-only & $0.233$ & $0.311$ & $0.277$ & $0.106$ \\
 & KL, mixed & $0.188$ & $0.076$ & $0.040$ & $0.014$ \\
\midrule
Harmful, AdvTok & No fine-tuning & $0.391$ & $0.690$ & $0.709$ & $0.583$ \\
 & KL, benign-only & $0.212$ & $0.339$ & $0.350$ & $0.133$ \\
 & KL, mixed & $0.173$ & $0.074$ & $0.039$ & $0.013$ \\
\midrule
Benign, random & No fine-tuning & $0.358$ & $0.322$ & $0.354$ & $0.346$ \\
 & KL, benign-only & $0.189$ & $0.129$ & $0.131$ & $0.048$ \\
 & KL, mixed & $0.195$ & $0.133$ & $0.139$ & $0.054$ \\
\bottomrule
\end{tabular}
\caption{Normalized residual distance $\|h^{\ell}(\boldt) - h^{\ell}(\boldt_c)\|^2 / \|h^{\ell}(\boldt_c)\|^2$ at the final prompt token, on OLMo-2-7B, averaged over 200 held-out AdvBench prompts or 200 held-out Alpaca prompts. Random encodings are four MDD draws per prompt. AdvTok encodings are those selected against the untrained model. Each arm is measured with its own model. All $95\%$ bootstrap half-widths over prompts are below $0.045$.}
\label{tab:residual-distance}
\end{table}

\section{Full-sequence StochasTok}\label{app.sec:stochastok}

StochasTok, as proposed for pretraining, is applied to the full sequence. It retokenizes the instruction and the answer at the same rate and trains on the retokenized answer tokens, leaving only the chat scaffolding canonical. We train six such arms at increasing $\alpha$ on our data mixture and evaluate each with the protocol of Section~\ref{subsec:asr} (Table~\ref{tab:stochastok-reversal}). The $\alpha = 0.3$ arm is the StochasTok baseline of Table~\ref{tab:attacks}.

Greedy AdvTok success falls monotonically with $\alpha$, from $\val{0.415}$ at $\alpha = 0$ to $\val{0.045}$ at $\alpha = 5.5$, and no arm produces an unsafe response to a canonically tokenized harmful prompt. Sampling at $T{=}1$ shows the same decrease, from $\val{0.576}$ at $\alpha = 0.1$ to $\val{0.079}$ at $\alpha = 5.5$. At high $\alpha$, full-sequence StochasTok is therefore as robust to AdvTok as context-only consistency training.

The difference lies in what the models write. At $\alpha = 1.7$ and $2.7$ they emit non-canonical tokens at per-token rates of $\val{0.84}$ and $\val{0.92}$, and at $\alpha = 5.5$ at $\val{0.42}$, against $\val{0.0001}$ at $\alpha = 0$. This is the behavior Proposition~\ref{prop:target} predicts for a model trained on retokenized targets, whose optimum reproduces the tokenizer's distribution over encodings. Context-only training keeps the output canonical. We did not measure benchmark accuracy for the arms above $\alpha = 0.3$.

At StochasTok's own depth the two approaches are close. The context-only arm of Appendix~\ref{app.sec:ablations}, which draws StochasTok-sampled encodings of the prompt under the KL objective and leaves the continuation canonical, reaches $\val{0.165}$ at greedy decoding, against $\val{0.205}$ for full-sequence StochasTok at $\alpha = 0.3$.

\begin{table}[htbp]
\centering
\small
\begin{tabular}{lcccc}
\toprule
$\alpha$ & Train inflation & AdvTok & Canonical unsafe & Non-canonical emission \\
\midrule
$0.0$ & $1.000\times$ & 0.415 & 0.000 & 0.0001 \\
$0.1$ & $1.087\times$ & 0.370 & 0.000 & 0.0172 \\
$0.3$ & $1.243\times$ & 0.205 & 0.000 & 0.0586 \\
$1.7$ & $2.074\times$ & 0.115 & 0.000 & 0.8368 \\
$2.7$ & $2.495\times$ & 0.080 & 0.000 & 0.9230 \\
$5.5$ & $3.309\times$ & 0.045 & 0.000 & 0.4218 \\
\bottomrule
\end{tabular}
\caption{Full-sequence StochasTok at increasing $\alpha$, greedy
decoding, 200 held-out AdvBench prompts, judged by Llama-Guard-3-8B.
Non-canonical emission is the per-token rate at which the model's own
output departs from the canonical tokenization of the text it decodes to.}
\label{tab:stochastok-reversal}
\end{table}

\section{Over-refusal}\label{app.sec:overrefusal}

Table~\ref{tab:overrefusal} reports refusal rates on canonical inputs
for nineteen arms, at training seed $42$. Benign refusal is measured on
100 held-out instructions, and XSTest-safe refusal on the 250 safe
prompts of XSTest \citep{rottger2024xstest}, which superficially
resemble unsafe requests. Responses are sampled with the model's chat
template, four per prompt at temperature 1 with 192 max new tokens,
each prompt contributing the mean refusal over its four samples.
Refusal is detected with the keyword-prefix judge, whose precision and
arm-dependent recall are characterized in Appendix~\ref{app.sec:judges};
intervals are half-widths of $\val{95}\%$ cluster bootstraps over prompts.
Under canonical tokenization every arm we re-evaluated refuses at least
$\val{0.97}$ of the 200 held-out AdvBench prompts, so we do not tabulate it.

The consistency arms leave benign refusal statistically unchanged
and move XSTest-safe refusal by at most $\val{+0.061}$ absolute (the 8-token
prefix arm; the output-KL and continuation-supervised arms sit at
$\val{+0.005}$ to $\val{+0.046}$). Over three training seeds, XSTest-safe
refusal averages $\val{0.068}$ for output KL and $\val{0.064}$ for
single-position representation matching at all layers. Context-perturbed SFT on dataset answers refuses $\val{0.235}$ of the XSTest-safe prompts at seed $42$, and $\val{0.092}$ and $\val{0.110}$ at two further seeds. Single-sided mixtures deform the refusal
boundary instead of extending it. The harmful-only arm refuses
indiscriminately, declining $\val{0.465}$ of benign instructions and
$\val{0.614}$ of XSTest-safe prompts.

\begin{table}[htbp]
\centering
\small
\begin{tabular}{lcc}
\toprule
Model & Benign refusal & XSTest-safe refusal \\
      & ($n{=}100$) & ($n{=}250$) \\
\midrule
No fine-tuning & $0.030 \pm 0.029$ & $0.065 \pm 0.024$ \\
\hdashline
Self-distillation, canonical only & $0.022 \pm 0.022$ & $0.102 \pm 0.029$ \\
Context-perturbed SFT, dataset answers & $0.052 \pm 0.035$ & $0.235 \pm 0.039$ \\
\hdashline
Output, cross-entropy, uniform & $0.037 \pm 0.029$ & $0.081 \pm 0.024$ \\
Output, cross-entropy, CVaR & $0.022 \pm 0.022$ & $0.085 \pm 0.027$ \\
\hdashline
Output, KL & $0.028 \pm 0.021$ & $0.097 \pm 0.027$ \\
Output KL, benign-only & $0.025 \pm 0.020$ & $0.045 \pm 0.017$ \\
Output KL, harmful-only & $0.465 \pm 0.065$ & $0.614 \pm 0.040$ \\
Output KL, final prompt position & $0.020 \pm 0.024$ & $0.057 \pm 0.020$ \\
\hdashline
Representation, single-position, final layer & $0.035 \pm 0.031$ & $0.090 \pm 0.031$ \\
Representation, single-position, multi-layer & $0.033 \pm 0.031$ & $0.048 \pm 0.021$ \\
Representation, single-position, all layers & $0.030 \pm 0.030$ & $0.079 \pm 0.030$ \\
Representation, continuation-supervised, final layer & $0.025 \pm 0.029$ & $0.111 \pm 0.031$ \\
Representation, continuation-supervised, multi-layer & $0.018 \pm 0.020$ & $0.070 \pm 0.025$ \\
\hdashline
Output KL, 8-token prefix & $0.015 \pm 0.015$ & $0.126 \pm 0.031$ \\
\hdashline
Multi-tokenization LoRA, identical copies & $0.022 \pm 0.026$ & $0.086 \pm 0.025$ \\
Multi-tokenization LoRA, length-ratio & $0.033 \pm 0.026$ & $0.095 \pm 0.027$ \\
Multi-tokenization LoRA, uniform & $0.015 \pm 0.014$ & $0.090 \pm 0.026$ \\
\hdashline
StochasTok, $\alpha{=}0.3$ & $0.028 \pm 0.024$ & $0.108 \pm 0.029$ \\
\bottomrule
\end{tabular}
\caption{Refusal on canonical inputs at training seed $42$. Benign refusal is measured on 100 held-out
instructions and XSTest-safe refusal on the 250 safe prompts of XSTest
\citep{rottger2024xstest}, written to resemble unsafe requests.
Responses are sampled with the chat template, four per prompt at
temperature 1, scored as the per-prompt mean. Refusal is detected by the
keyword-prefix judge characterized in Appendix~\ref{app.sec:judges}. Values
are the rate $\pm$ half the width of a $95\%$ cluster bootstrap over
prompts.}
\label{tab:overrefusal}
\end{table}

\section{Attack settings}\label{app.sec:attacks}

Every attack in Tables~\ref{tab:attacks} and~\ref{tab:rev-attacks} is the authors' reference implementation at its published settings, with the template correction described below, run separately against each model, so each model faces inputs found against itself. The prompts are the same 200 held-out AdvBench requests for every attack and every arm, formatted with the model's chat template, with no system prompt.

\noindent\textbf{AdvTok} \citep{geh2025adversarial} runs $100$ greedy search iterations, sampling $8$ tokenizations per iteration from the segmentation lattice. Responses are $256$ new tokens, decoded greedily or sampled eight times at $T{=}1$.

\noindent\textbf{BEAST} \citep{sadasivan2024fast} appends an adversarial suffix of $40$ tokens found by beam search with $k_1{=}k_2{=}15$.

\noindent\textbf{Charmer} \citep{abadrocamora2024charmer} edits the prompt itself: up to $k{=}20$ character substitutions chosen over $5$ candidate positions per step for $50$ steps.

BEAST and Charmer are scored under a separate pipeline from AdvTok: responses are $128$ new tokens, decoded greedily or sampled three times at $T{=}1$. All three are judged unsafe by Llama-Guard-3-8B (Appendix~\ref{app.sec:judges}); the reported rate is the fraction of prompts judged unsafe at $T{=}0$ and the fraction of samples at $T{=}1$, with a $\val{95}\%$ cluster bootstrap over prompts. For rows that average several training seeds, the interval is computed on the per-prompt mean across seeds. Search budgets differ by attack, so rates are comparable within a column of Table~\ref{tab:rev-attacks} and not across attacks. The repr-next arm supervises the residual stream at the final prompt position at every layer.

\noindent\textbf{Evaluation fix.} The submitted version evaluated AdvTok through the upstream implementation's OLMo code path, which had two defects: responses were generated without an assistant turn, and the search omitted a newline from the chat template. Both affect evaluation only. Training used a separate template builder, which we verified to match the model's chat template token for token, and the sensitivity sweeps of Appendix~\ref{app.sec:splittability} and Appendix~\ref{app.sec:domains} never passed through the defective path. We corrected the template and re-ran every attack against every checkpoint. All attack results in the main text and in Appendix~\ref{app.sec:rebuttal-results} use the corrected evaluation. Absolute rates changed; for example, greedy AdvTok success on the untrained model is $0.400$, where the submitted version reported $0.565$. Appendix tables that have not been re-evaluated are marked as such in their captions. In the corrected runs the AdvTok search starts from a random encoding, and a single search per prompt is shared between $T{=}0$ and $T{=}1$.

\section{Judges}\label{app.sec:judges}

Attack success is judged by Llama-Guard-3-8B. Table~\ref{tab:harmbench} repeats the main AdvTok results under the HarmBench classifier, on one training seed. HarmBench gives a lower rate in every cell, and every trained arm remains below the untrained model under both judges. On 56 hand-labeled responses that Llama-Guard judged unsafe, its precision is $\val{0.696}$ (39 harmful compliance, 9 off-request, 7 refusal, 1 degenerate). A second sample of 50 responses, drawn where our instruments disagreed, gives a precision of $\val{0.667}$ and a recall of 26 of 26. Both samples have a single annotator and were generated before the evaluation fix of Appendix~\ref{app.sec:attacks}, so they describe the earlier responses. Degenerate responses are handled as described in Appendix~\ref{app.subsec:rev-judge}.

Refusal is detected by a fixed keyword-prefix matcher applied to the
first 200 characters of the lowercased response. Its precision is
high on a hand-labeled sample ($5/5$; the false-positive rate is
otherwise unmeasured). Its recall varies by arm, with missed-refusal
rates between $\val{0.070}$ and $\val{0.305}$ at eight character substitutions,
highest exactly where reported refusal is lowest, so refusal gaps
between arms are partly instrument-driven and we lean on orderings
rather than magnitudes.

\begin{table}[tb]
\centering
\small
\setlength{\tabcolsep}{3.4pt}
\newcommand{\cj}[1]{{\scriptsize$\,\pm$#1}}
\caption{AdvTok attack success on the $200$ held-out AdvBench prompts under two
judges: Llama-Guard-3-8B \citep{inan2023llama}, which every other table uses,
and the HarmBench classifier \citep{mazeika2024harmbench}. $T$ is the decoding
temperature. Values are rates $\pm$ half the width of a $95\%$ bootstrap interval over prompts, at training seed $42$ for every row, so the KL and
repr-next rows differ from the seed means of Table~\ref{tab:attacks}. The CE
row is the uniform-weighting run of Appendix~\ref{app.sec:cvar}. HarmBench reads
lower than Llama-Guard in every cell, and it leaves the ordering of the arms
unchanged apart from self-distillation and StochasTok, which it inverts.}
\label{tab:harmbench}
\begin{tabular}{@{}lcccc@{}}
\toprule
 & \multicolumn{2}{c}{Llama-Guard-3-8B} & \multicolumn{2}{c}{HarmBench} \\
\cmidrule(lr){2-3}\cmidrule(l){4-5}
 & $T=0$ & $T=1$ & $T=0$ & $T=1$ \\
\midrule
    No fine-tuning & $0.400$\cj{.067} & $0.458$\cj{.052} & $0.235$\cj{.060} & $0.219$\cj{.037} \\
    Self-distillation & $0.325$\cj{.065} & $0.426$\cj{.046} & $0.130$\cj{.047} & $0.157$\cj{.028} \\
    StochasTok, $\alpha{=}0.3$ & $0.205$\cj{.055} & $0.336$\cj{.050} & $0.155$\cj{.050} & $0.163$\cj{.033} \\
\midrule
    Context-perturbed SFT & $0.020$\cj{.018} & $0.062$\cj{.026} & $0.010$\cj{.013} & $0.024$\cj{.014} \\
    Consistency, CE (uniform) & $0.045$\cj{.027} & $0.072$\cj{.023} & $0.015$\cj{.018} & $0.018$\cj{.010} \\
    Consistency, KL & $0.010$\cj{.013} & $0.019$\cj{.015} & $0.005$\cj{.007} & $0.007$\cj{.006} \\
    Consistency, repr-next & $0.010$\cj{.013} & $0.009$\cj{.011} & $0.005$\cj{.007} & $0.005$\cj{.006} \\
\bottomrule
\end{tabular}
\end{table}

\section{Probes}\label{app.sec:probes}

The frozen-probe results of Section~\ref{sec:probes} use the corrected attack and the 63 held-out harmful prompts. The six topics are cyber/malware (58 of the 200 harmful prompts), fraud/theft (36), misinformation/deception (20), weapons/explosives (15), violence/self-harm (13) and a residual class (58); on the held-out prompts the largest class holds $\val{0.35}$. The refit-probe table, the drift analysis, the predictor table, the directed-search results and the steering controls below were computed on the attack encodings of the submitted version and have not been re-evaluated.

The probes are fit once, on the untrained
model's canonical activations, and applied to every arm without
refitting. Table~\ref{tab:probe-refit} reports the complementary
instrument. For each arm we fit a fresh logistic probe on that arm's
own canonical activations at layer 32, with the same prompt-level
70/30 split and seed, and evaluate it on canonical tokenizations, on
constructed one- and two-split edits, on tokenizations drawn uniformly
from the MDD, and on the attack encodings of the shared condition. Under the tokenizations selected against the untrained model, the frozen probe reads $\val{1.000}$ on the KL model, $\val{0.968}$ on CE and $\val{1.000}$ on repr-next, and $\val{0.444}$ on the benign-only KL model, a gain of $\val{0.127}$ over the untrained model.

Two facts justify the frozen instrument in the main text. Refitting
saturates on trained arms, with seven of ten rows at $\val{0.95}$ or above
under attack, so comparisons that separate the arms under frozen
transfer, such as $\val{0.968}$ against $\val{0.698}$ at matched layer and
condition, compress here to $\val{1.000}$
against $\val{0.952}$. Refitting also recovers nothing on the untrained
model, which reads $\val{0.397}$ under attack with either instrument; the
suppression of Section~\ref{sec:probes} is therefore a genuine absence of
the signal rather than a rotation a fixed probe would miss. Where the
instruments disagree the gap is modest and layer-specific: the
single-position KL arm reads $\val{0.952}$ refit against $\val{0.873}$ frozen
under the shared attack at layer 32, and identically at layers 16 and
24. Accuracy falls from the split edits to the deeper uniform MDD samples
for the untrained model, canonical SFT, and the benign-only arm, while
every mixed-trained arm holds ceiling on both.

\noindent\textbf{Drift and crack rate.} Pooling $1{,}600$ prompt-arm pairs, the crack rate
rises monotonically with layer-32 residual drift within every arm, from
$\val{0.02}$--$\val{0.07}$ below drift $\val{0.01}$ to $\val{0.33}$--$\val{1.0}$ above $\val{0.3}$. The arms
differ chiefly in where their drift mass sits: the mixed-trained arms place
$150$--$166$ prompts in the lowest bin, while the untrained model places
none below $\val{0.03}$ and $142$ between $0.3$ and $1.0$. There is no common absolute
threshold; the untrained model cracks \emph{less} than treated arms at
matched drift in the middle bins, so tolerance scales with an arm's typical
drift rather than with drift itself. Consistency training compresses the
drift of harmful-prompt attack encodings into the regime where the frozen refusal
probe still classifies correctly. We report binned rates rather than a
fitted arm effect because four of the six cells above drift $\val{0.3}$ contain
$n \le 8$ pairs and arm is nearly collinear with bin.

\noindent\textbf{Prospective predictors.} Neither the
model's refusal margin at the canonical tokenization nor the frozen probe's
margin there predicts attack outcome (Table~\ref{tab:probe-auc}). The
reading-sensitivity summaries of Section~\ref{app.sec:splittability}
predict weakly and are subsumed by the probe margin under the attacked
tokenization, which predicts strongly but is measured on the encoding the
search selected. It confirms the suppression mechanism per prompt; it does not
forecast vulnerability.

\begin{table}[h]
\centering
\small
\begin{tabular}{lc}
\toprule
Predictor of attack outcome & AUC \\
\midrule
Refusal margin, canonical & $0.485$ \\
Probe margin, canonical & $0.507$--$0.532$ \\
$\Delta_i$ summaries (median, spread, worst, CVaR-25\%) & $0.62$--$0.64$ \\
Probe margin, attacked tokenization (post hoc) & $0.85$ \\
Joint over probe margin alone & $+0.001$ \\
\bottomrule
\end{tabular}
\caption{Predicting per-prompt attack outcome on the untrained model.
The post-hoc predictor is measured on the encoding the search
selected.}
\label{tab:probe-auc}
\end{table}

\begin{table}[htbp]
\centering
\small
\begin{tabular}{lcccc}
\toprule
Model & Canonical & Splits & MDD-uniform & Adversarial \\   
\midrule
No fine-tuning & $1.000$ & $0.987$ & $0.812$ & $0.397$ \\
Canonical SFT & $0.992$ & $0.979$ & $0.825$ & $0.429$ \\
Output, KL & $1.000$ & $1.000$ & $1.000$ & $1.000$ \\
Output, cross-entropy & $1.000$ & $1.000$ & $0.998$ & $1.000$ \\
Output KL, benign-only & $1.000$ & $0.996$ & $0.885$ & $0.667$ \\
Output KL, final prompt position & $1.000$ & $1.000$ & $0.990$ & $0.952$ \\
Representation, single-position, final layer & $1.000$ & $1.000$ & $1.000$ & $1.000$ \\
Representation, single-position, multi-layer & $1.000$ & $1.000$ & $0.997$ & $0.984$ \\
Representation, continuation-supervised, final layer & $1.000$ & $1.000$ & $0.998$ & $1.000$ \\
Representation, continuation-supervised, multi-layer & $1.000$ & $1.000$ & $0.998$ & $1.000$ \\
\bottomrule
\end{tabular}
\caption{Refit-per-arm probe accuracy at layer 32. A logistic probe is
fit for each arm on that arm's own canonical activations, with the
prompt-level split of Section~\ref{sec:probes}, and evaluated on
canonical tokenizations, on constructed split edits, on uniform MDD
retokenizations, and on the attack encodings selected against the
untrained model.}
\label{tab:probe-refit}
\end{table}

\noindent\textbf{Directed searches.} This is what its objective requires. Fluent
compliance presupposes that the instruction is still read, so the search is
constrained to preserve comprehension while it is free, and pressured, to
destroy the feature that refusal depends on. The probes confirm both halves
of the selection at the representation level. On the prompts that crack, the
dissociation is sharpest: refusal-probe accuracy is $\val{0.167}$ while topic decodability is $\val{0.583}$. A search directed at the topic probe suppresses topic decodability from the same site on $\val{0.735}$ of the prompts it reads correctly at canonical ($36/49$), at $\val{1.78}\times$ inflation against the compliance attack's $\val{2.69}\times$. The refusal-probe margin stays positive on $\val{0.905}$ of all $63$ encodings, and refusal declines from $\val{0.730}$ to $\val{0.587}$ (paired $\val{-0.143}$), indistinguishable from the depth-matched control ($\val{-0.048}$). Each search removes the feature its objective targets; the untargeted probe stays near its depth-matched baseline in both cases.
\noindent\textbf{Probe protocol.} Both probes use a prompt-level split, so no
prompt contributes tokenizations to both halves, and both are frozen
thereafter. A two-layer probe is fit identically, to distinguish information
that is absent from information that is present but not linearly decodable. We
evaluate two adversarial conditions. Under the \emph{shared} condition every
model receives the tokenizations selected by the attack against the untrained
model, which asks how each model represents one fixed attack. Under the
\emph{own} condition each model receives the tokenizations selected against
itself, the condition matching its reported attack success. Probes are read at
layers $\{8,16,24,32\}$; we report layer 32 unless stated. Refitting a probe
per model saturates and conceals the differences between arms: the
final-prompt-position arm reads $\val{0.952}$ under a refit probe and $\val{0.873}$ under
the frozen one, on the shared condition at layer 32.

\noindent\textbf{Steering controls.} A random direction of matched norm
restores nothing ($\val{0.73}$--$\val{0.85}$ unsafe at every scale), and the effect is
dose-dependent ($\val{0.150}$, $\val{0.761}$, $\val{1.000}$ refusal at $\val{0.5}\times$, $\val{1}\times$,
$\val{2}\times$). The direction itself is not prompt-specific: at the calibrated
scale it also induces refusal on $\val{0.550}$ of benign prompts, consistent with the
indiscriminate refusal steering of \citet{arditi2024refusal}, so the
intervention demonstrates mechanism rather than supplying a defense.

\section{Additional models, seeds and attacks}\label{app.sec:rebuttal-results}
{\color{revblue}
This appendix repeats the main experiments on Llama-3.2-3B-Instruct and gemma-3-4b-it and over additional training seeds of OLMo-2-7B-Instruct. Unless stated otherwise, attack cells give the share of responses that Llama-Guard-3-8B judges unsafe at $T{=}0$ / $T{=}1$, with degenerate responses excluded (Section~\ref{app.subsec:rev-judge}). Among the fine-tuned arms of a model, \textbf{bold} marks the best value in a column and \textit{italic} the second best, with no second best when the best is tied; the untrained model is not ranked. In the tables, ``Repr., last pos.'' is repr-next and ``Repr., last pos.\ + answer'' is repr-cont. On OLMo-2-7B the KL rows average two training seeds, the repr-next rows three, and the repr-cont.\ rows use one. These searches were run separately from those of Table~\ref{tab:attacks}.

\subsection{Attacks on three model families}\label{app.subsec:rev-attacks}

Consistency training lowers the success of adversarial tokenization on all three models (Table~\ref{tab:rev-attacks}). Transfer to the attacks that change the text is less uniform: representation matching at the last prompt position lowers BEAST success on every model, whereas the KL objective and representation matching with the answer positions raise it on gemma and OLMo.

\begin{table}[H]
\centering
\small\color{revblue}
\setlength{\tabcolsep}{5pt}
\newcommand{\ci}[1]{{\scriptsize$\,\pm$#1}}
\caption{Consistency training reduces adversarial tokenization success on every model, while its effect on BEAST and Charmer depends on the objective. Cells are attack success at $T{=}0$ / $T{=}1$.}\label{tab:rev-attacks}
\begin{tabular}{@{}lcccc@{}}
\toprule
 & No attack & AdvTok & BEAST & Charmer \\
\midrule
\multicolumn{5}{@{}l}{\textit{Llama-3.2-3B-Instruct}} \\
\quad No fine-tuning & 0.010 / 0.022 & 0.320 / 0.394 & 0.410 / 0.530 & 0.480 / 0.428 \\
\quad StochasTok & 0.005 / \textit{0.018} & 0.050 / 0.110 & 0.265 / 0.548 & 0.200 / \textit{0.320} \\
\quad Consistency, KL & \textbf{0.000} / 0.019 & \textbf{0.005} / \textit{0.022} & \textit{0.225} / 0.410 & \textbf{0.170} / 0.328 \\
\quad Consistency, Repr., last pos. & 0.005 / 0.020 & 0.015 / 0.030 & \textbf{0.170} / \textbf{0.212} & 0.365 / 0.358 \\
\quad Consistency, Repr., last pos.\ + answer & \textbf{0.000} / \textbf{0.009} & \textbf{0.005} / \textbf{0.013} & 0.295 / \textit{0.328} & \textit{0.180} / \textbf{0.152} \\
\midrule
\multicolumn{5}{@{}l}{\textit{gemma-3-4b-it}} \\
\quad No fine-tuning & 0.030 / 0.042 & 0.385 / 0.372 & 0.465 / 0.400 & 0.085 / 0.078 \\
\quad StochasTok & 0.085 / 0.059 & 0.195 / 0.377 & \textit{0.480} / \textit{0.555} & 0.075 / 0.095 \\
\quad Consistency, KL & \textbf{0.000} / \textbf{0.011} & \textit{0.095} / 0.106 & 0.735 / 0.603 & \textit{0.035} / \textit{0.033} \\
\quad Consistency, Repr., last pos. & \textit{0.005} / \textbf{0.011} & \textbf{0.040} / \textbf{0.051} & \textbf{0.245} / \textbf{0.335} & 0.155 / 0.187 \\
\quad Consistency, Repr., last pos.\ + answer & 0.010 / 0.014 & \textit{0.095} / \textit{0.091} & 0.760 / 0.625 & \textbf{0.025} / \textbf{0.008} \\
\midrule
\multicolumn{5}{@{}l}{\textit{OLMo-2-7B-Instruct, mean over training seeds}} \\
\quad No fine-tuning & -- & 0.400 / 0.458 & 0.655 / 0.543 & 0.500 / 0.483 \\
\quad StochasTok & -- & 0.185 / 0.331 & \textbf{0.400} / \textit{0.460} & 0.585 / 0.608 \\
\quad Consistency, KL & -- & \textbf{0.028} / 0.050 & 0.816 / 0.674 & \textit{0.416} / 0.442 \\
\quad Consistency, Repr., last pos. & -- & \textit{0.032} / \textbf{0.030} & \textit{0.455} / \textbf{0.420} & 0.467 / \textit{0.379} \\
\quad Consistency, Repr., last pos.\ + answer & -- & 0.040 / \textit{0.041} & 0.745 / 0.605 & \textbf{0.365} / \textbf{0.325} \\
\bottomrule
\end{tabular}
\end{table}

\subsection{Jailbreak wrappers and prefill}\label{app.subsec:rev-wrappers}

A retokenization search on top of a jailbreak wrapper or of a prefill placed in the user turn makes both far more effective against the untrained models. Consistency training removes most of this increase, and the representation objectives remove nearly all of it (Table~\ref{tab:rev-wrappers}). These evaluations were not run on gemma.

\begin{table}[H]
\centering
\small\color{revblue}
\setlength{\tabcolsep}{5pt}
\newcommand{\ci}[1]{{\scriptsize$\,\pm$#1}}
\caption{Consistency training removes most of the gain that a retokenization search adds to jailbreak wrappers and to user-turn prefill. Cells are attack success at $T{=}0$ / $T{=}1$.}\label{tab:rev-wrappers}
\begin{tabular}{@{}lcccc@{}}
\toprule
 & \multicolumn{2}{c}{Wrapper} & \multicolumn{2}{c}{Prefill (user turn)} \\
\cmidrule(lr){2-3}\cmidrule(lr){4-5}
 & alone & + retok.\ search & alone & + retok.\ search \\
\midrule
\multicolumn{5}{@{}l}{\textit{Llama-3.2-3B-Instruct}} \\
\quad No fine-tuning & 0.040 / 0.050 & 0.250 / 0.264 & 0.060 / 0.033 & 0.140 / 0.223 \\
\quad StochasTok & \textbf{0.000} / 0.025 & 0.040 / 0.185 & \textbf{0.000} / \textit{0.007} & 0.080 / 0.102 \\
\quad Consistency, KL & \textbf{0.000} / 0.037 & 0.050 / 0.102 & 0.020 / 0.030 & 0.020 / 0.030 \\
\quad Consistency, Repr., last pos. & 0.010 / \textit{0.020} & \textit{0.030} / \textit{0.046} & \textbf{0.000} / 0.013 & \textbf{0.000} / \textit{0.010} \\
\quad Consistency, Repr., last pos.\ + answer & 0.010 / \textbf{0.010} & \textbf{0.020} / \textbf{0.035} & \textbf{0.000} / \textbf{0.005} & \textbf{0.000} / \textbf{0.005} \\
\midrule
\multicolumn{5}{@{}l}{\textit{OLMo-2-7B-Instruct, mean over training seeds}} \\
\quad No fine-tuning & 0.120 / 0.166 & 0.680 / 0.701 & 0.000 / 0.005 & 0.860 / 0.823 \\
\quad StochasTok & \textbf{0.040} / \textit{0.094} & 0.290 / 0.494 & \textbf{0.000} / \textit{0.030} & 0.420 / 0.547 \\
\quad Consistency, KL & 0.160 / 0.193 & 0.237 / 0.307 & 0.020 / 0.054 & 0.093 / 0.185 \\
\quad Consistency, Repr., last pos. & \textit{0.063} / \textbf{0.067} & \textbf{0.120} / \textbf{0.134} & \textbf{0.000} / \textbf{0.002} & \textbf{0.007} / \textbf{0.013} \\
\quad Consistency, Repr., last pos.\ + answer & 0.070 / 0.107 & \textit{0.150} / \textit{0.182} & \textbf{0.000} / 0.040 & \textit{0.020} / \textit{0.055} \\
\bottomrule
\end{tabular}
\end{table}

\subsection{Degenerate responses and the judge}\label{app.subsec:rev-judge}

Some fine-tuned models occasionally produce degenerate text, such as a single repeated token. We call a response degenerate if it has fewer than three words, if fewer than $20\%$ of its words are distinct, or if it is long and compresses to less than a fifth of its length. Llama-Guard judges $73\%$ of such responses unsafe, against $16\%$ of the others, although they contain no harmful content, so we do not count them as attack successes. The exclusion matters for few cells, mostly those of representation matching at the last prompt position (Table~\ref{tab:rev-degenerate}).

\begin{table}[H]
\centering
\small\color{revblue}
\setlength{\tabcolsep}{5pt}
\newcommand{\ci}[1]{{\scriptsize$\,\pm$#1}}
\caption{Excluding degenerate responses changes the attack success of few cells. The table lists every cell in which at least $5\%$ of the responses are degenerate, at $T{=}0$ / $T{=}1$; bold marks a decrease of at least $0.05$.}\label{tab:rev-degenerate}
\begin{tabular}{@{}lllccc@{}}
\toprule
Model & Arm & Attack & As judged & Excluded & Degenerate \\
\midrule
Llama-3.2-3B & Repr., last pos. & No attack & 0.190 / 0.123 & \textbf{0.005} / \textbf{0.020} & 24\% / 13\% \\
Llama-3.2-3B & Repr., last pos. & AdvTok & 0.170 / 0.101 & \textbf{0.015} / \textbf{0.030} & 22\% / 10\% \\
Llama-3.2-3B & Repr., last pos. & Charmer & 0.575 / 0.402 & \textbf{0.365} / 0.358 & 24\% / 5\% \\
gemma-3-4b & KL & BEAST & 0.775 / 0.715 & 0.735 / \textbf{0.603} & 4\% / 12\% \\
gemma-3-4b & Repr., last pos. & AdvTok & 0.080 / 0.083 & 0.040 / 0.051 & 6\% / 5\% \\
gemma-3-4b & Repr., last pos. & BEAST & 0.640 / 0.533 & \textbf{0.245} / \textbf{0.335} & 41\% / 21\% \\
gemma-3-4b & Repr., last pos. & Charmer & 0.240 / 0.213 & \textbf{0.155} / 0.187 & 9\% / 3\% \\
gemma-3-4b & Repr., last pos.\ + answer & BEAST & 0.825 / 0.763 & \textbf{0.760} / \textbf{0.625} & 7\% / 15\% \\
\bottomrule
\end{tabular}
\end{table}

\subsection{Capability and over-refusal}\label{app.subsec:rev-capability}

Every fine-tuned arm loses some benchmark accuracy. The loss is smallest on OLMo-2-7B and larger on the two smaller models, where it depends strongly on the objective. Repr-cont.\ loses least on Llama and gemma. StochasTok and repr-next lose most of gemma's GSM8K accuracy, and repr-next loses $0.293$ MMLU on Llama (Table~\ref{tab:rev-capability}). On Llama every fine-tuned arm refuses more of the safe XSTest prompts than the untrained model, and StochasTok most; on gemma only StochasTok does (Table~\ref{tab:rev-overrefusal}).

\begin{table}[H]
\centering
\small\color{revblue}
\setlength{\tabcolsep}{5pt}
\newcommand{\ci}[1]{{\scriptsize$\,\pm$#1}}
\caption{Benchmark accuracy after fine-tuning. The loss is smallest on OLMo-2-7B and larger, and objective-dependent, on the smaller models. The first row of each model is absolute; the other rows are paired differences from it with $95\%$ confidence intervals.}
\label{tab:rev-capability}
\begin{tabular}{@{}lccc@{}}
\toprule
 & MMLU (5-shot) & GSM8K (8-shot CoT) & IFEval (0-shot) \\
\midrule
\multicolumn{4}{@{}l}{\textit{Llama-3.2-3B-Instruct}} \\
\quad No fine-tuning & 0.575 & 0.759 & 0.693 \\
\quad StochasTok & $-$0.068\ci{.008} & $-$0.257\ci{.029} & $-$0.322\ci{.047} \\
\quad Consistency, KL & \textit{$-$0.024}\ci{.006} & $-$0.075\ci{.023} & $-$0.216\ci{.043} \\
\quad Consistency, Repr., last pos. & $-$0.293\ci{.010} & \textbf{$-$0.024}\ci{.023} & \textit{$-$0.189}\ci{.044} \\
\quad Consistency, Repr., last pos.\ + answer & \textbf{$-$0.005}\ci{.005} & \textbf{$-$0.024}\ci{.021} & \textbf{$-$0.133}\ci{.039} \\
\midrule
\multicolumn{4}{@{}l}{\textit{OLMo-2-7B-Instruct}} \\
\quad No fine-tuning & 0.597 & 0.823 & 0.682 \\
\quad Canonical-only self-distillation & $-$0.020\ci{.006} & $-$0.042\ci{.020} & $-$0.067\ci{.037} \\
\quad StochasTok & $-$0.094\ci{.008} & $-$0.173\ci{.025} & $-$0.176\ci{.047} \\
\quad Context-perturbed SFT & $-$0.058\ci{.008} & $-$0.128\ci{.024} & $-$0.185\ci{.044} \\
\quad Consistency, CE (CVaR) & \textbf{$-$0.006}\ci{.005} & $-$0.064\ci{.022} & $-$0.059\ci{.039} \\
\quad Consistency, CE (uniform) & \textit{$-$0.011}\ci{.005} & \textbf{$-$0.015}\ci{.018} & \textit{$-$0.054}\ci{.040} \\
\quad Consistency, KL & $-$0.039\ci{.007} & \textit{$-$0.033}\ci{.020} & \textbf{$-$0.046}\ci{.039} \\
\quad Consistency, Repr., last pos. & $-$0.042\ci{.007} & $-$0.047\ci{.021} & $-$0.124\ci{.039} \\
\quad Consistency, Repr., last pos.\ + answer & $-$0.012\ci{.005} & $-$0.045\ci{.020} & $-$0.105\ci{.038} \\
\midrule
\multicolumn{4}{@{}l}{\textit{gemma-3-4b-it}} \\
\quad No fine-tuning & 0.532 & 0.745 & 0.730 \\
\quad StochasTok & \textbf{$+$0.004}\ci{.009} & $-$0.539\ci{.029} & $-$0.397\ci{.047} \\
\quad Consistency, KL & $-$0.104\ci{.008} & \textit{$-$0.092}\ci{.026} & \textit{$-$0.227}\ci{.044} \\
\quad Consistency, Repr., last pos. & $-$0.095\ci{.008} & $-$0.541\ci{.029} & $-$0.440\ci{.047} \\
\quad Consistency, Repr., last pos.\ + answer & \textit{$-$0.011}\ci{.007} & \textbf{$-$0.077}\ci{.024} & \textbf{$-$0.137}\ci{.039} \\
\bottomrule
\end{tabular}
\end{table}

\begin{table}[H]
\centering
\small\color{revblue}
\setlength{\tabcolsep}{5pt}
\newcommand{\ci}[1]{{\scriptsize$\,\pm$#1}}
\caption{StochasTok raises refusal of safe prompts on both models, and the consistency arms do so only on Llama. Cells are keyword refusal rates on canonical prompts with $95\%$ confidence intervals; the last column is the share of degenerate answers on XSTest-safe at $T{=}0$, and arms above $5\%$ are not ranked.}\label{tab:rev-overrefusal}
\begin{tabular}{@{}lcccc@{}}
\toprule
 & \multicolumn{2}{c}{XSTest-safe} & Alpaca & Degenerate \\
\cmidrule(lr){2-3}
 & $T{=}0$ & $T{=}1$ & $T{=}0$ & \\
\midrule
\multicolumn{5}{@{}l}{\textit{Llama-3.2-3B-Instruct}} \\
\quad No fine-tuning & 0.060\ci{.045} & 0.062\ci{.040} & 0.000 & 0\% \\
\quad StochasTok & 0.270\ci{.090} & 0.226\ci{.064} & 0.040 & 0\% \\
\quad Consistency, KL & 0.190\ci{.075} & \textit{0.163}\ci{.052} & 0.000 & 0\% \\
\quad Consistency, Repr., last pos. & \textbf{0.110}\ci{.060} & \textbf{0.141}\ci{.053} & 0.000 & 0\% \\
\quad Consistency, Repr., last pos.\ + answer & \textit{0.180}\ci{.075} & 0.188\ci{.066} & 0.000 & 0\% \\
\midrule
\multicolumn{5}{@{}l}{\textit{gemma-3-4b-it}} \\
\quad No fine-tuning & 0.070\ci{.050} & 0.065\ci{.048} & 0.000 & 0\% \\
\quad StochasTok & 0.250\ci{.080} & 0.233\ci{.052} & 0.040 & 0\% \\
\quad Consistency, KL & \textbf{0.040}\ci{.035} & \textit{0.052}\ci{.028} & 0.000 & 0\% \\
\quad Consistency, Repr., last pos. & 0.030\ci{.035} & 0.029\ci{.033} & 0.000 & 29\% \\
\quad Consistency, Repr., last pos.\ + answer & \textbf{0.040}\ci{.035} & \textbf{0.046}\ci{.039} & 0.000 & 0\% \\
\bottomrule
\end{tabular}
\end{table}

\subsection{Harmful share of the training mixture}\label{app.subsec:rev-mixture}

For AdvTok, the three harmful shares give values within the range across training seeds at the main share ($0.015$ to $0.060$ at greedy decoding), so we detect no effect. BEAST and Charmer success are lower with more harmful prompts, with one training seed per share (Table~\ref{tab:rev-mixture}).

\begin{table}[H]
\centering
\small\color{revblue}
\setlength{\tabcolsep}{5pt}
\newcommand{\ci}[1]{{\scriptsize$\,\pm$#1}}
\caption{Attack success at three harmful shares of the training mixture. OLMo-2-7B-Instruct, representation matching at the last position; cells are attack success at $T{=}0$ / $T{=}1$.}\label{tab:rev-mixture}
\begin{tabular}{@{}lccc@{}}
\toprule
Harmful share & AdvTok & BEAST & Charmer \\
\midrule
0.10 & 0.035 / 0.054 & 0.495 / 0.398 & 0.575 / 0.512 \\
0.286 (\textit{main experiments}) & \textbf{0.015} / \textbf{0.014} & \textit{0.360} / \textit{0.345} & \textit{0.430} / \textit{0.353} \\
0.50 & \textbf{0.015} / \textit{0.021} & \textbf{0.295} / \textbf{0.305} & \textbf{0.405} / \textbf{0.340} \\
\bottomrule
\end{tabular}
\end{table}

}

\end{document}